%% file: main.tex
\documentclass[10pt,twocolumn,letterpaper]{article}

\usepackage{cvpr}              
\usepackage[accsupp]{axessibility}
\usepackage{graphicx}
\usepackage{amsmath}
\usepackage{amssymb}
\usepackage{booktabs}

\usepackage{array, booktabs}
\usepackage{comment}
\usepackage{algorithm}
\usepackage{algorithmic}

\usepackage{bbding}
\usepackage{pifont}
\usepackage{wasysym}
\usepackage{amssymb}

\newcommand\blfootnote[1]{%
  \begingroup
  \renewcommand\thefootnote{}\footnote{#1}%
  \addtocounter{footnote}{-1}%
  \endgroup
}

\usepackage[pagebackref,breaklinks,colorlinks]{hyperref}

\usepackage[capitalize]{cleveref}
\crefname{section}{Sec.}{Secs.}
\Crefname{section}{Section}{Sections}
\Crefname{table}{Table}{Tables}
\crefname{table}{Tab.}{Tabs.}

\def\cvprPaperID{11561} 
\def\confName{CVPR}
\def\confYear{2022}

\input{preamble}
\begin{document}
\input{sec/0_metadata}

\maketitle

\blfootnote{$^*$ denotes equally contributed.}

\begin{abstract}
%
%
We propose a new 3D spatial understanding task for 3D question answering (3D-QA). In the 3D-QA task, models receive visual information from the entire 3D scene of a rich RGB-D indoor scan and answer given textual questions about the 3D scene.
Unlike the 2D-question answering of visual question answering, the conventional 2D-QA models suffer from problems with spatial understanding of object alignment and directions and fail in object localization from the textual questions in 3D-QA. We propose a baseline model for 3D-QA, called the ScanQA\footnote{\url{https://github.com/ATR-DBI/ScanQA}}, which learns a fused descriptor from 3D object proposals and encoded sentence embeddings. This learned descriptor correlates language expressions with the underlying geometric features of the 3D scan and facilitates the regression of 3D bounding boxes to determine the described objects in textual questions. We collected human-edited question--answer pairs with free-form answers grounded in 3D objects in each 3D scene. Our new ScanQA dataset contains over 41k question--answer pairs from 800 indoor scenes obtained from the ScanNet dataset. To the best of our knowledge, ScanQA is the first large-scale effort to perform object-grounded question answering in 3D environments.
\end{abstract}

\input{sec/introduction.tex}
\input{sec/related_work}

\input{sec/dataset.tex}
\input{sec/baseline_models}

\input{sec/experiments}

\input{sec/conclusion}


{
    \small
    \bibliographystyle{ieee_fullname}
    \bibliography{egbib}
}

\input{sec/X_supplementary}


\end{document}

%% file: preamble.tex
\usepackage{overpic}
\usepackage{enumitem} 
\usepackage{overpic} 
\usepackage{color}

\definecolor{turquoise}{cmyk}{0.65,0,0.1,0.3}
\definecolor{purple}{rgb}{0.65,0,0.65}
\definecolor{dark_green}{rgb}{0, 0.5, 0}
\definecolor{orange}{rgb}{0.8, 0.6, 0.2}
\definecolor{red}{rgb}{0.8, 0.2, 0.2}
\definecolor{darkred}{rgb}{0.6, 0.1, 0.05}
\definecolor{blueish}{rgb}{0.0, 0.3, .6}
\definecolor{light_gray}{rgb}{0.7, 0.7, .7}
\definecolor{pink}{rgb}{1, 0, 1}
\definecolor{greyblue}{rgb}{0.25, 0.25, 1}

\usepackage{blindtext}

\renewcommand{\paragraph}[1]{\vspace{1em}\noindent\textbf{#1}.}

%% file: sec/0_metadata.tex
\title{ScanQA: 3D Question Answering for Spatial Scene Understanding}

\author{
    Daichi Azuma$^*$\\
    Kyoto University
    \and
    Taiki Miyanishi$^*$\\
    ATR,~~RIKEN AIP
    \and
    Shuhei Kurita$^*$\\
    RIKEN AIP,~~JST PRESTO
    \and
    Motoaki Kawanabe\\
    ATR,~~RIKEN AIP
    
}

%% file: sec/introduction.tex
\section{Introduction}
\begin{figure}[t]
\begin{center}
\includegraphics[scale=0.42,clip]{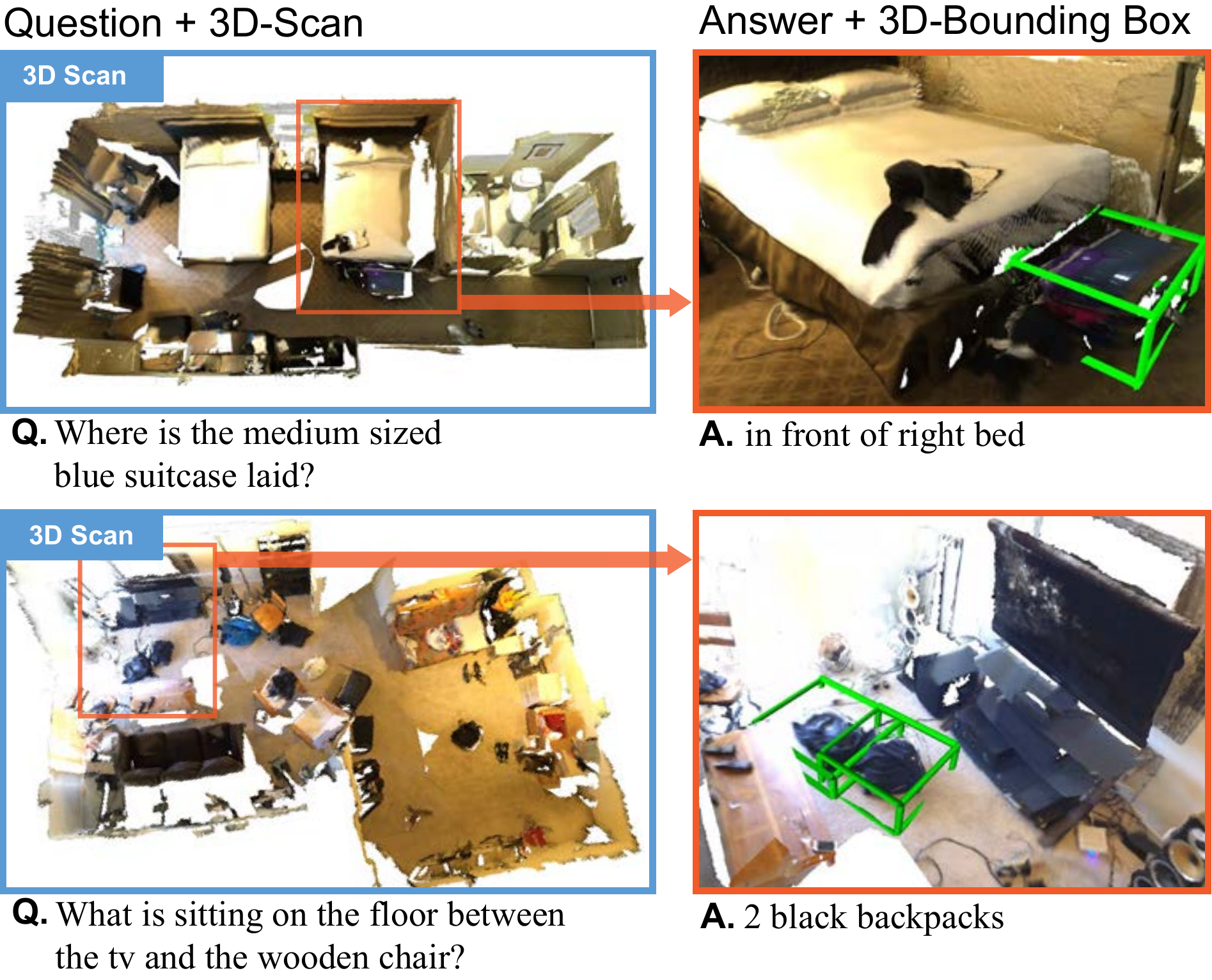}
\end{center}
\vspace{-0.4cm}
\caption{
We introduce the new task of question answering for 3D modeling.
Given inputs of an entire 3D modeling and a linguistic question, models predict an answer phrase and the corresponding 3D-bounding boxes.
}
\label{fig:teaser}
\vspace{-4mm}
\end{figure}
In recent years, significant advances have been achieved in vision-and-language tasks and datasets, and several new datasets have been created to develop models that understand textual expressions, such as captions or questions, which are grounded in two-dimensional (2D) images, such as image captioning\cite{mscoco,sharma2018conceptual}, understanding referring expressions~\cite{referitgame,refcoco}, image region and phrase correspondence~\cite{flickr30k}, and visual question answering (VQA)~\cite{vqa_iccv2015,Goyal_2017_CVPR,hudson2018gqa}.
VQA is successful in grasping object features visualized in 2D frames. However, when we develop models that understand the spatial information of 3D scenes, such as ``What is between the table and TV set?'' or ``Where is the suitcase located?'', the existing models based on 2D images have several challenges in accurately understanding the 3D world. For example, 2D images lack an accurate sense of the relative directions and distances in the 3D scenes, i.e., the stereoscopic attribute-perception problem. Some objects are hidden by other objects when they overlap, i.e., the occlusion problem.  When multiple images are used in 2D-image-based question answering models, such models often encounter difficulties in tracking and recognizing whether some objects are the same object between images, i.e., the object localization and identification problem.


Currently, 3D spatial-understanding models can be developed for the 3D object localization task of ScanRefer~\cite{chen2020scanrefer},  the dialog-based localization of ReferIt3D~\cite{achlioptas2020referit_3d}, and the 3D object captioning task of Scan2Cap~\cite{chen2021scan2cap}.
Embodied question answering ~\cite{embodiedqa,eqa_matterport,mt_eqa} is an important task for navigating agents in a 3D scene. We consider that 3D spatial understanding datasets contribute to developing models that comprehend the embodied 3D scene and ask and answer questions about the 3D environment as humans do.
However, unlike their 2D image counterparts, question answering datasets on 3D environmental annotations are still limited in terms of dataset size and question variety because existing datasets often rely on template-based question--answer collections.


In this paper, we propose a 3D question answering (3D-QA) task that uses 3D spatial information instead of 2D images to comprehend real-world information through the question answering form.
In the 3D-QA task, models answer a question for a 3D scene as well as the object localization described in the question.
We present the overview of the task in Fig.~\ref{fig:teaser}.
This 3D-QA task setting is reasonable when external sensors or mobile robots collect sufficient visual information to construct a 3D scene before the QA task. We assume that this is plausible when the model can use the preliminarily captured visual information from the 3D scene because of prior navigation in the scene, such as vision-and-language navigation~\cite{vln}. This task is also applicable to real-world services that use preliminarily extracted 3D scenes, such as interactive virtual room-viewing services or searching in indoor scenes.


For the 3D-QA task, we developed a novel ScanQA dataset based on RGB-D scans of an indoor scene and annotations derived from the ScanNet dataset~\cite{dai2017scannet}. We automatically generated questions from the object captions of ScanRefer~\cite{chen2020scanrefer} using question generation models. However, these auto-generated questions included many invalid questions; therefore, we filtered the invalid questions and refined them if necessary. We collected free-form answers and object annotations from humans using a newly developed interactive 3D scene viewer. In total, we gathered 41k question--answer pairs with 32k unique questions. 
We propose a 3D-QA model with textual and 3D scene encoding and several baseline models, including 2D image models (2D-QA), a combination of 3D object localization models ~\cite{chen2020scanrefer,Qi_2019_ICCV}, and a question answering model~\cite{Yu_2019_CVPR}.
We confirmed that the ScanQA model outperformed the baseline models in most evaluations, including exact matching and image captioning metrics in the proposed ScanQA dataset.

%% file: sec/related_work.tex
\section{Related work}

\input{sec/table_dataset.tex}
The 3D-QA task is similar to existing visual question answering and 3D embodied question answering.
We place our task as a spatial comprehension of the entire 3D scene given linguistic questions.

\subsection{Visual Question Answering}
Visual question answering (VQA) is a task in which models are given a 2D image and a question about its content. They are expected to provide an appropriate answer. Question answering in 2D images was proposed by Malinowski \textit{et al.}~\cite{malinowski2014}, and various inference methods~\cite{vqa_iccv2015, Anderson_2018_CVPR, Yu_2019_CVPR} have since been proposed.
One of the best VQA methods is Oscar~\cite{li2020oscar}, which uses Mask R-CNN to infer a solution by considering the relationship between individual objects in the image.
In addition, Jang \textit{et al.}~\cite{jang-IJCV-2019} proposed a question answering method that considers more detailed vision and motion information using video. ClipBERT~\cite{lei2021less} improved its accuracy by dividing a video into clips and reasoning from them individually.
VQA $360^{\circ}$~\cite{vqa360} is a task for answering questions about a $360^{\circ}$ image. Although VQA $360^{\circ}$ contributes to understanding the 3D scene, the available information is limited compared with the ScanQA dataset. Our dataset also includes the object identification task, which is different from the existing VQA $360^{\circ}$ dataset.

\subsection{3D Object Localization with Language}
ScanRefer~\cite{chen2020scanrefer} localizes an object referred to in a free-form text description.
The ScanRefer model identifies a 3D-bounding box for an object given the input description.
This dataset is based on 800 scenes derived from the ScanNet dataset~\cite{dai2017scannet}.
In addition to answering questions on 3D scenes, the 3D-QA task also includes an object localization task for objects that appear in the question answering.
Unlike in the ScanRefer task, the objects in ScanQA object localization can be multiple because multiple objects can appear in a single question.

\subsection{Question Answering in 3D scenes}
Unlike 2D-QA, for which many datasets have been proposed,
3D question answering datasets are still limited. We notice that the existing question answering tasks on 3D scenes have interactive forms.
The interactive QA dataset (IQUAD)~\cite{gordon2018iqa} on AI2THOR~\cite{ai2thor} enables model agents to interact with objects in a scene to determine the answer to a question. Embodied question answering (EQA)~\cite{embodiedqa,eqa_matterport,mt_eqa} is a combination of visual question answering and navigation such as vision-and-language navigation tasks ~\cite{vln,roomnav,touchdown,alfred} and models~\cite{speaker-follower,glgp}. In the original EQA dataset~\cite{embodiedqa}, the embodied model agent receives a question  such as ``What color is the car?'' and navigate to the object described in the question in House 3D~\cite{roomnav}. MP3D-EQA~\cite{eqa_matterport}  is a photorealistic embodied QA for Matterport 3D scans~\cite{Matterport3D}. MT-EQA~\cite{mt_eqa} is a multitarget variation of EQA.
We summarize the relation of these datasets with ScanQA in Table~\ref{table:datasets}.
Unlike these datasets, the ScanQA dataset is not created from fixed templates and hence includes more natural and a significantly larger number of unique questions, as discussed in Sec.~\ref{sec:dataset_stat}.

%% file: sec/table_dataset.tex

\renewcommand{\arraystretch}{0.93}

\begin{table*}[t]
\begin{center}
	\footnotesize\begin{tabular}{lcccccc}
        \toprule
3D-QA Datasets       & Type        & Question Collection & Answer Collection & Environment& Photorealistic & \# 3D Scenes \\
\midrule
IQUAD                & Interactive & Template-based & Template-based   & AI2THOR       & No  & 30 rooms \\
EQA                  & Navigation  & Template-based & Template-based   & House3D       & No  & 588 scenes\\
MP3D-EQA             & Navigation  & Template-based & Template-based   & Matterport 3D & Yes & 144 floors\\
MT-EQA               & Navigation  & Template-based & Template-based   & House3D       & No  & 588 scenes\\
\midrule
ScanQA dataset      & 3D Scan     & AutoGen+HumanEdit & Human & ScanNet & Yes & 800 rooms \\
\bottomrule
	\end{tabular}
	\vspace{-0.2cm}
    \caption{
        Comparison of 3D question-answering datasets. 
    }
    \label{table:datasets}
\end{center}
\vspace{-0.6cm}
\end{table*}

\if[]
\begin{table}[t]
\begin{center}
	\footnotesize\begin{tabular}{lcccccc}
        \toprule
3D-QA        & Type        & Question  & Answer  & Environment& Photoreal & \# 3D Scenes \\
\midrule
IQUAD                & Interactive & Template & Template   & AI2THOR       & No  & 30 rooms \\
EQA                  & Navigation  & Template & Template   & House3D       & No  & 588 scenes\\
MP3D-EQA             & Navigation  & Template & Template   & Matterport 3D & Yes & 144 floors\\
MT-EQA               & Navigation  & Template & Template   & House3D       & No  & 588 scenes\\
\midrule
ScanQA       & 3D Scan     & AutoGen & Human & ScanNet & Yes & 800 rooms \\
\bottomrule
	\end{tabular}
	\vspace{-0.2cm}
    \caption{
        Comparison of 3D question-answering datasets. 
    }
\end{center}
\vspace{-0.5cm}
\end{table}
\fi

%% file: sec/dataset.tex
\section{ScanQA Dataset}
We hereby define the 3D-QA task and describe the collection of the corresponding dataset.

\subsection{3D-QA Task}
As illustrated in Fig.~\ref{fig:teaser}, a 3D-QA task requires models to answer a question when given all the information of a 3D scene.
Here, models use the 3D spatial information, such as RGB-D scans or point cloud data. 
We also require models to specify the 3D-bounding boxes of objects that are related to this question answering. This prevents models from answering questions by relying on the textual priors of the trained questions without examining the scene.
However, unlike the ScanRefer dataset, we do not require models to target one described object for each question. This is because multiple objects can be used to answer certain questions. For example, the question ``What color is the chairs around the table?'' is related to multiple objects. This question is also answerable as long as the chairs around the unique table in the scene have the same color. 
In such scenarios, we require models to answer the question addressing multiple 3D-bounding boxes.

\subsection{Question-Answer Collection}
\begin{figure}[t]
\begin{center}
\includegraphics[scale=0.94,clip]{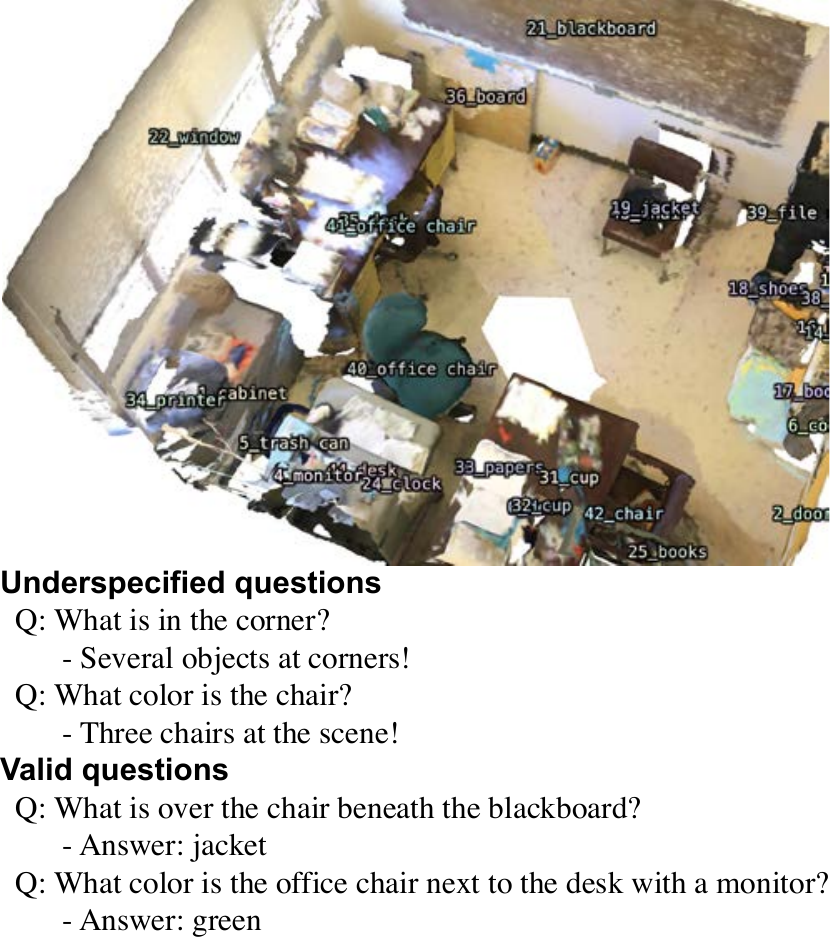}
\end{center}
\vspace{-0.4cm}
\caption{
%
Underspecified and valid questions for an office room scene.
We presented scenes with object IDs and names to MTurk workers for the dataset collection.
}
\label{fig:dataset_example}
\vspace{-0.5cm}
\end{figure}

The ScanQA dataset was created using multiple phrases, including automatic QA generation~\cite{Yang_2021_ICCV,alberti_2019-synthetic,chan_2019_emlnlp-w,lopez_2021_pricai}, question filtering, question editing, and answer collection. 
First, we automatically generated question--answer pairs from the referring expressions to identify objects in 3D scenes obtained from the ScanRefer dataset~\cite{chen2020scanrefer}. We applied the question-and-answer generation model based on a T5-base model~\cite{2020t5} trained on a text-based question answering dataset~\cite{rajpurkar_2016_squad}~\footnote{We used the weights available at \url{https://huggingface.co/valhalla/t5-base-qa-qg-hl}.} for ScanRefer captions and obtained the seed questions.
However, these autogenerated question--answer pairs included many inadequate questions, such as those that are underspecified or not grounded in the scene, as presented in Fig.~\ref{fig:dataset_example}.
Auto-generated questions also include easy questions that can be answered with common sense.
Therefore, we decided to remove such questions as much as possible.
We also did not include auto-generated answers in the final dataset because it was not clear that they were grounded in scenes.
Then, we applied filtering and editing to the seed questions with basic rules and human editing in addition to the answer collection via Amazon Mechanical Turk (MTurk).
For human editing and answer collection, we developed an interactive visualization website for each 3D scene that enables workers to interact with the 3D scene and check the object names and IDs if they are available. Following the ScanRefer dataset, we attached the object names and IDs for the objects in the 3D scene. We embedded this site into the MTurk task page (Fig.~\ref{fig:dataset_example}).

The filtering and editing of the seed questions were conducted as follows.
First, we filtered the inadequate questions from the auto-generated seed questions using basic rules.
Subsequently, we asked the workers to classify the remaining questions into four classes: \textit{valid}, \textit{too easy}, \textit{unanswerable}, and \textit{unclear} questions. Each question was evaluated by at least three workers. We selected questions in which two or more workers were marked as valid for the next phrase of the editing and answer collection process.
In the editing and answer collection, we first presented the filtered questions to workers and requested them to rewrite the questions themselves if they were inadequate for the scene before writing a free-form answers.
Multiple answers are collected when necessary.
We also collected the object IDs that were used in the question to identify the object in the scene in this phrase. See SM~\ref{sec:supp_mturk} for details.

\begin{table}[t]
\begin{center}
	\footnotesize\begin{tabular}{lccc}
        \toprule
Split       & \# Question & \# Unique Question & \# 3D Scenes \\
\midrule
Train            & 25,563 & 20,546 & 562 \\
Val              & 4,675  &  4,306 & 71 \\
Test w/ objects  & 4,976  &  4,552 & 70 \\
Test w/o objects & 6,149  &  5,484 & 97 \\
\midrule
Total            & 41,363 & 32,337 & 800 \\
\bottomrule
	\end{tabular}
	\vspace{-0.2cm}
    \caption{
        ScanQA dataset statistics.
    }
    \label{table:dataset_stat}
\end{center}
\vspace{-0.8cm}
\end{table}

\subsection{Dataset Statistics}
\label{sec:dataset_stat}
\begin{figure}[t]
\begin{center}
\includegraphics[scale=0.72,clip]{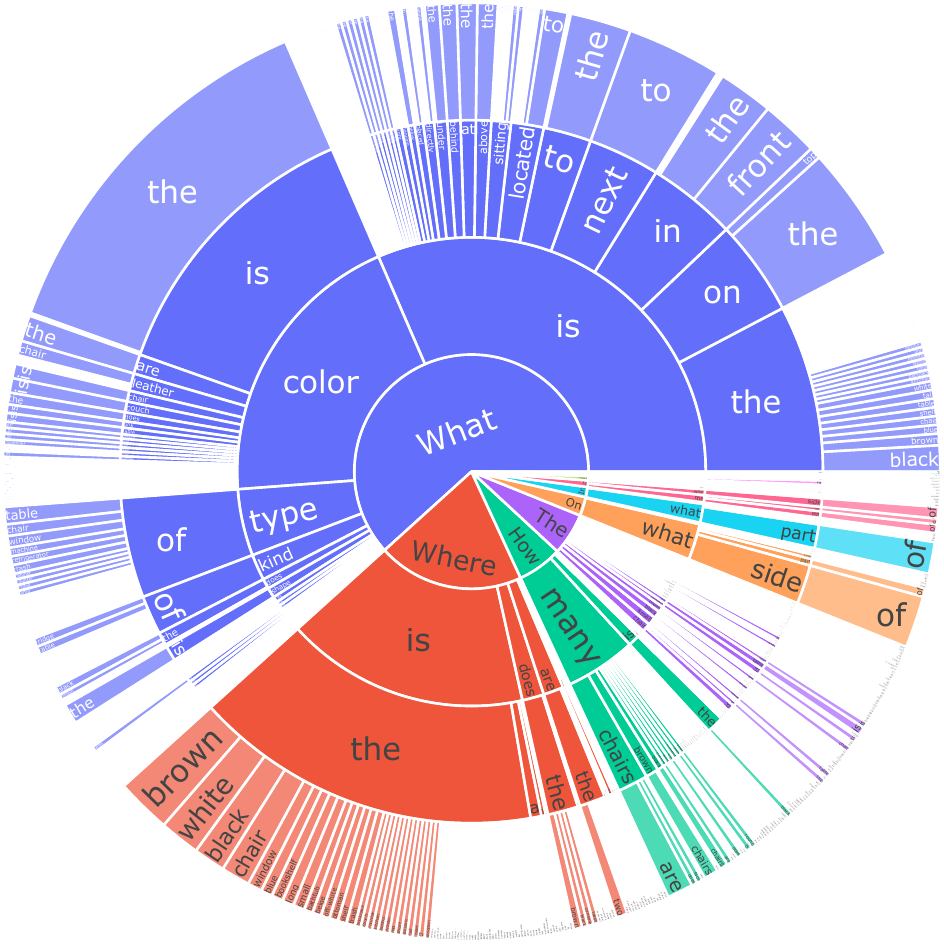}
\end{center}
\vspace{-0.6cm}
\caption{
The distribution of the question types by the beginning of the question writing.
}
\label{fig:dataset_circle}
\vspace{-0.4cm}
\end{figure}

We collected 41,363 questions and 58,191 answers, including 32,337 unique questions and 16,999 unique answers.
Table~\ref{table:dataset_stat} presents the statistics of the ScanQA dataset.
This dataset is an order of magnitude larger than existing embodied question-answering datasets in terms of both question size and variation.
For example, the EQA dataset~\cite{embodiedqa} contains 4,246 questions, consisting of 147 unique questions in its training set.
The EQA-MP3D dataset~\cite{eqa_matterport} contains 767 questions consisting of 174 unique questions in its training set.
Considering that our dataset contains not only question--answer pairs but also 3D object localization annotations, we assume that this is the largest dataset to specify the nature of objects in 3D scenes with the question answering form.
The distribution of the questions based on their first word is shown in Fig.~\ref{fig:dataset_circle}.
We collected various types of questions  through question auto-generation and editing by humans.

We followed the training, validation, and test set splits used in ScanRefer. However, as the object IDs for the test set of ScanRefer are not publicly available, we further split the validation set of ScanRefer into two-holds as the validation set and test set with object annotations in the ScanQA dataset. Therefore, the ScanQA dataset includes two test sets with and without object annotations.
We collected at least two answers for each question in the validation and two test sets to evaluate the free-form answers.
As our dataset includes the question type of ``Where is,'' writing expressions for answers can vary. Therefore, we adopted evaluation metrics for image captioning in addition to an exact match to the annotated answers in the evaluation.

%% file: sec/baseline_models.tex
\section{ScanQA Model}~\label{sec:proposed_method}
We introduce the baseline model of ScanQA for the 3D-QA task.
The 3D-QA is formalized as follows:
given inputs of the point cloud $p \in \mathcal{P}$ and question $q \in \mathcal{Q}$ about the 3D scene, the 3D-QA model aims to output $\hat{a}$ that semantically matches true answer $a^*$.

\noindent \textbf{3D feature representation.}
We primarily use the input point cloud $p$ consisting of point coordinates $c\in \mathbb{R}^3$ in the 3D space for 3D representation.
Following previous 3D and language research~\cite{chen2020scanrefer,chen2021scan2cap}, we use additional point features such as the height of the point, 
colors, normals, and multiview image features~\cite{dai20183dmv}
that project 2D appearance features to the point cloud.
We use these combined point features as 3D features $r \in \mathbb{R}^{135}$.

\begin{figure}
 \centering
 \includegraphics[keepaspectratio, scale=0.78]{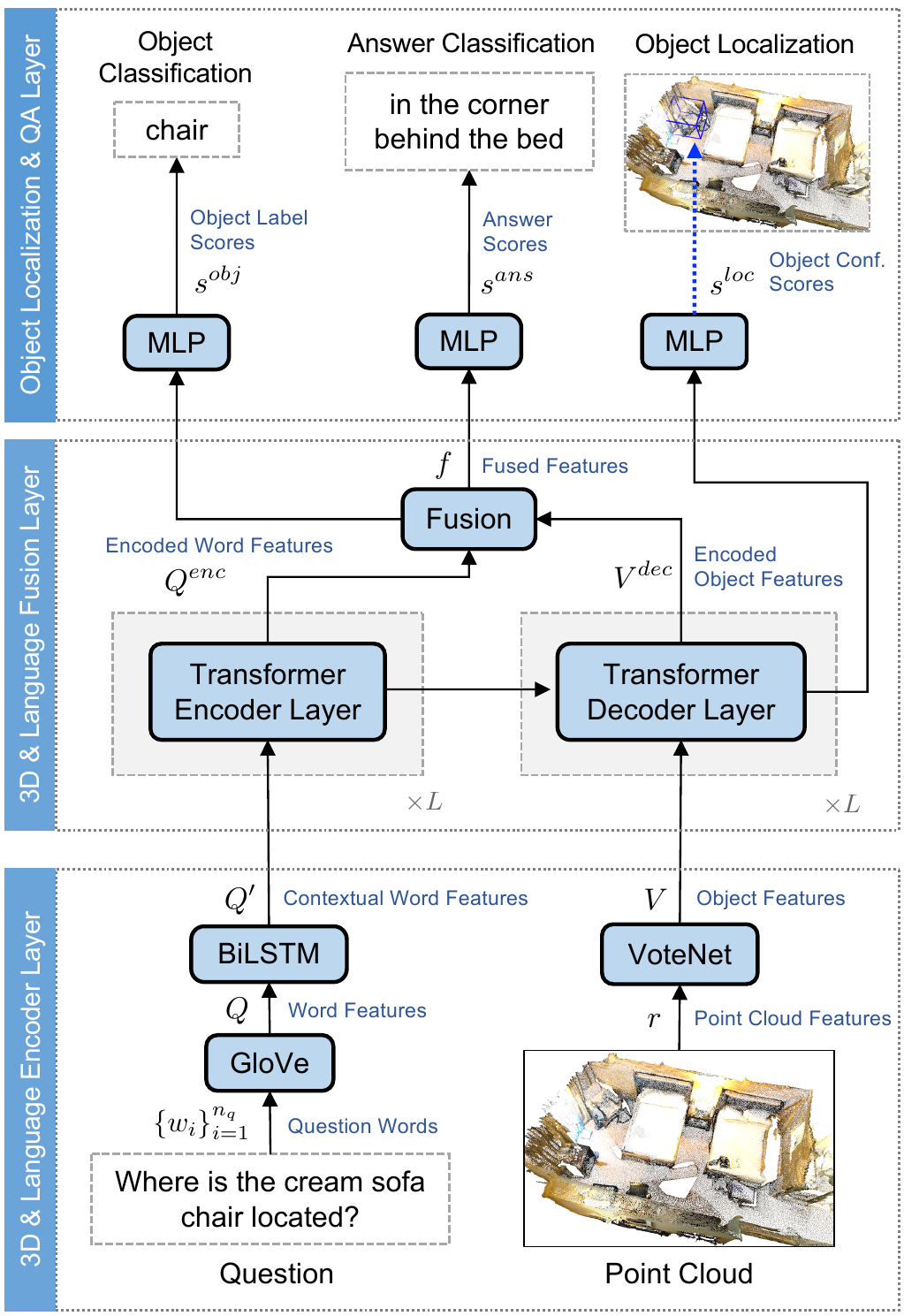} 
 \vspace{-0.2cm}
 \caption{ScanQA model for answering 3D environments. Given a point cloud and RGB frame sequence that capture indoor scenes, the QA model outputs a corresponding answer by fusing 3D and language information
 through three layers: the 3D scene and language encoder layer, fusion layer, and classification layers.}
 \label{fig:scanqa_ovierview}
 \vspace{-0.5cm} 
\end{figure}

\noindent \textbf{Overview of network architecture.}
To solve the 3D-QA task, we developed a ScanQA model consisting of a 3D \& language encoder, 3D \& language fusion, and object localization \& QA layers.
An overview of the proposed ScanQA network is presented in~Fig.~\ref{fig:scanqa_ovierview}.
The 3D \& language encoder layer transforms the question into contextualized word representations and point clouds into object proposals. The 3D \& language fusion layer combines multiple 3D object features guided by language information using transformer-based encoder and decoder layers~\cite{NIPS2017_3f5ee243,Yu_2019_CVPR}. 
The object localization \& QA layer estimates the target object box and object labels and predicts answers associated with questions and scene content.

\noindent \textbf{3D \& language encoder layers.}
This layer encodes the question words $\{w_i\}^{n_q}_{i=1}$ using GloVe~\cite{pennington2014glove}, and we obtain word representation $Q \in \mathbb{R}^{n_{q}\times 300}$, where $n_q$ is the number of words in question, and feeds them into a one-layer bidirectional long short-term memory (biLSTM)~\cite{hochreiter_1997_neuralcomp} for word sequence modeling.
We project a series of output states from the LSTM
using a nonlinear layer with GELUs~\cite{Hendrycks_2016_arxiv} activation to obtain the contextualized word representation $Q' \in \mathbb{R}^{n_q\times d}$, where $d$ is the hidden size of the biLSTM (set to 256).
In addition, this layer detects objects in a scene based on point cloud features $r \in \mathbb{R}^{135}$ using VoteNet~\cite{Qi_2019_ICCV}, which uses PointNet++~\cite{qi_2017_NIPS} as a backbone network.
We obtain the object proposals (object boxes) from VoteNet and project them using a nonlinear layer with GELUs activation to obtain the object proposal representation $V \in \mathbb{R}^{n_{v}\times d}$,
where $n_{v}$ is the number of object proposals 
(set to 256.)

\noindent \textbf{3D \& language fusion layer.}
Inspired by the architecture of deep modular co-attention networks of MCAN~\cite{Yu_2019_CVPR}, often 
used for VQA, we use transformer blocks~\cite{NIPS2017_3f5ee243} to represent the relationships between object proposals and between question words.
After feeding contextual question representation $Q'$ into 
a stack of $L$ (set to two) transformer encoder layers,
we obtain a deeply contextualized question representation $Q^{enc} \in \mathbb{R}^{n_q\times d}$.
In addition, we use transformer decoder layers to represent the features of object proposals related to the question words by using the final output of the transformer encoder as the decoder's keys and values.
We obtain a question-aware-object proposal representation $V^{dec}\in \mathbb{R}^{n_v\times d}$ after feeding the question and object proposal pairs into a stack of $L$ transformer decoder layers.
Subsequently, the final outputs of the transformer layers $Q^{enc}$ and $V^{dec}$ are fused by a fusion layer that uses
two-layer multi-layer perceptron (MLP) with an attention mechanism~\cite{luong_2015_emlnlp} (for details, see~\cite{Yu_2019_CVPR}). 
We obtain the fused feature $f \in \mathbb{R}^{d}$, which simultaneously represents a 3D scene and linguistic question information.

\noindent \textbf{Object localization \& QA layer.}
This layer consists of object localization, object classification, and answer classification modules.
Each module is described as follows.

\noindent \textit{\textbf{Object localization module}} 
aims to predict which of the proposed object boxes corresponds to the question. 
The question-aware-object proposal representation, $V^{dec}\in \mathbb{R}^{n_v\times d}$, is fed into a two-layer MLP to determine the likelihood of each object box being related to the question.
Following~\cite{chen2020scanrefer}, we compute the localization confidence $s^{loc} \in \mathbb{R}^{n_v}$ for the proposed $n_v$ object boxes with the cross-entropy (CE) loss to train this module.

\noindent \textit{\textbf{Object classification module}} 
predicts what objects are associated with a question.
Note that many questions do not contain target object names related to the answer, in contrast to a 3D localization task~\cite{chen2020scanrefer,Zhao_2021_ICCV,Yuan_2021_ICCV}.
We use the 3D and question-aware fused feature $f$ and feed it into a two-layer MLP
to predict 18 ScanNet benchmark classes.
We compute the object classification scores $s^{obj} \in \mathbb{R}^{18}$ with a softmax function and use the CE loss to train this module.

\noindent \textit{\textbf{Answer classification module}} 
predicts an answer corresponding the question and scene.
We project the fused feature $f$ into a vector $s^{ans} \in \mathbb{R}^{n_a}$
for the $n_a$ answer candidates in the training set.
To consider multiple answers, we compute final scores
with the binary cross-entropy (BCE) loss function to train the module.

\noindent \textbf{Loss function.}
We use a loss function similar to ScanRefer~\cite{chen2020scanrefer}, such as the localization loss $\mathcal{L}_{loc}$ of the object localization module, object detection loss $\mathcal{L}_{det}$ of VoteNet~\cite{Qi_2019_ICCV}, and object classification loss $\mathcal{L}_{obj}$ of the object classification module.
To answer the 3D scene content, we additionally use the answer loss $\mathcal{L}_{ans}$ of the answer classification module. We set the final loss as a simple linear combination of these losses, computed as $\mathcal{L} = \mathcal{L}_{ans} + \mathcal{L}_{obj} + \mathcal{L}_{loc} + \mathcal{L}_{det}$.

%% file: sec/experiments.tex
\begin{table*}[t]
\begin{center}
	\footnotesize\begin{tabular}{lcccccccccc}
        \toprule
Model                & EM@1 & EM@10 & BLEU-1 & BLEU-2 & BLEU-3 & BLEU-4 & ROUGE & METEOR & CIDEr & SPICE \\
\midrule
\textbf{Test w/ objects} \\
RandomImage+MCAN  & 22.31 & 53.11 & 26.66 & 18.49 & 16.16 & 14.26 & 31.27 & 12.13 & 60.37 & 9.05\\
VoteNet+MCAN 	 &  19.71 & 50.76 & 29.46 & 17.23 & 10.33 & 6.08 & 30.97 & 12.07 & 58.23 & 10.44 \\
ScanRefer+MCAN (pipeline) & 17.52 & 49.92 & 19.17 & 10.66 & 0.00 & 0.00 & 24.40 & 9.38 & 44.25 & 6.24 \\
ScanRefer+MCAN (e2e) 	 &  20.56 & 52.35 & 27.85 & 17.27 & 11.88 & 7.46 & 30.68 & 11.97 & 57.36 & 10.58 \\
ScanQA 	 &  \textbf{23.45} & \textbf{56.51} & \textbf{31.56} & \textbf{21.39} & \textbf{15.87} & \textbf{12.04} & \textbf{34.34} & \textbf{13.55} & \textbf{67.29} & \textbf{11.99} \\
\midrule
OracleImage+MCAN   & 25.34 & 55.93 & 28.70 & 20.11 & 16.78 & 12.89 & 34.59 & 13.42 & 67.24 & 11.93\\
\midrule
\textbf{Test w/o objects} \\
RandomImage+MCAN  & 20.82 & 51.23 & 26.29 & 17.90 & 14.27 & 9.66 & 29.23 & 11.54 & 55.64 & 8.87 \\
VoteNet+MCAN 	 &  18.15 & 48.56 & 29.63 & 17.80 & 11.57 & 7.10 & 29.12 & 11.68 & 53.34 & 10.36 \\
ScanRefer+MCAN (pipeline) & 16.47 & 49.05 & 18.71 & 10.98 & 16.53 & 0.76 & 22.45 & 8.76 & 40.81 & 6.41 \\
ScanRefer+MCAN (e2e) 	 &  19.04 & 49.70 & 26.98 & 16.17 & 11.28 & 7.82 & 28.61 & 11.38 & 53.41 & 10.63 \\
ScanQA 	 & \textbf{20.90} & \textbf{54.11} & \textbf{30.68} & \textbf{21.20} & \textbf{15.81} & \textbf{10.75} & \textbf{31.09} & \textbf{12.59} & \textbf{60.24} & \textbf{11.29} \\
\bottomrule
	\end{tabular}
	\vspace{-0.2cm}
    \caption{
Performance comparison of question answering with image captioning metrics. \textbf{e2e} represents an end-to-end model.
    }
    \label{table:performance3D}
\end{center}
\vspace{-0.4cm}
\end{table*}

\begin{table*}[t]
\begin{center}
	\footnotesize\begin{tabular}{ccccccccccccc}
        \toprule

ANS & OBJ & LOC & EM@1 & EM@10 & BLEU-1 & BLEU-2 & BLEU-3 & BLEU-4 & ROUGE & METEOR & CIDEr & SPICE \\
\midrule
\multicolumn{3}{l}{\textbf{Test w/ objects}} \\
\checkmark &  & &  12.16 & 42.77 & 12.86 & 5.45 & 0.12 & 0.02 & 17.55 & 6.71 & 29.17 & 4.05 \\
\checkmark &  \checkmark & &  18.31 & 49.18 & 22.32 & 14.53 & 11.15 & 7.92 & 26.37 & 9.94 & 49.10 & 6.36 \\
\checkmark &  & \checkmark &  20.46 & 51.67 & 25.06 & 16.91 & 14.06 & 11.37 & 29.22 & 11.13 & 55.17 & 8.21 \\
\checkmark & \checkmark & \checkmark & \textbf{23.45} & \textbf{56.51} & \textbf{31.56} & \textbf{21.39} & \textbf{15.87} & \textbf{12.04} & \textbf{34.34} & \textbf{13.55} & \textbf{67.29} & \textbf{11.99} \\
\midrule
\multicolumn{3}{l}{\textbf{Test w/o objects}} \\
\checkmark & & &  10.78 & 39.44 & 11.94 & 5.02 & 0.12 & 0.02 & 15.34 & 5.91 & 25.51 & 3.51 \\
\checkmark &  \checkmark & &  16.23 & 46.30 & 21.37 & 13.49 & 10.71 & 7.64 & 23.63 & 9.10 & 43.21 & 6.13 \\
\checkmark &  & \checkmark &  18.12 & 49.60 & 25.12 & 17.58 & 14.68 & 10.23 & 26.50 & 10.43 & 49.93 & 8.16 \\
\checkmark & \checkmark & \checkmark &  \textbf{20.90} & \textbf{54.11} & \textbf{30.68} & \textbf{21.20} & \textbf{15.81} & \textbf{10.75} & \textbf{31.09} & \textbf{12.59} & \textbf{60.24} & \textbf{11.29} \\
\bottomrule
	\end{tabular}
	\vspace{-0.2cm}	
    \caption{
        Performance comparison of different experimental conditions of the ScanQA model.        
    }
    \label{table:component_ablation}
\end{center}
\vspace{-0.6cm}
\end{table*}

\begin{table}[t]
\begin{center}
	\footnotesize\begin{tabular}{lcc}
        \toprule
Model                & Acc@0.25 &  EM@10 \\
\midrule
\textbf{Test w/ objects} \\
ScanQA (xyz)  	 &  23.67 & 55.67 \\
ScanQA (xyz+rgb)  	 &  23.45 & 55.43 \\
ScanQA (xyz+rgb+normal)   	 &  23.35 & 54.50 \\
ScanQA (xyz+multiview)  	 &  25.40 & 55.51 \\
ScanQA (xyz+multiview+normal) 	 &  25.44 & 56.51 \\
\bottomrule
	\end{tabular}
	\vspace{-0.2cm}
    \caption{
        Feature ablation results
    }
    \label{table:feature_ablation}
\end{center}
\vspace{-0.7cm}
\end{table}

\section{Experiments}
\subsection{Experimental Setup}~\label{sec:exp_setup}
In this experimental setup, we referred to the experimental setup of existing studies on scene understanding of ScanNet~\cite{dai2017scannet}
through languages such as ScanRefer and Scan2Cap~\cite{chen2020scanrefer,chen2021scan2cap}.

\noindent \textbf{Data augmentation.} 
We applied data augmentation to our training data and applied rotation about all three axes using a random angle in $[-5^{\circ}, 5^{\circ}]$ and randomly translated the point cloud within 0.5 m in all directions. Because the ground alignment in ScanNet was incomplete, we rotated it on all axes (not just the top).

\noindent \textbf{Training.}
To train the ScanQA model, we used Adam~\cite{kingma2014adam}, a batch size of $16$, and an initial learning rate of $5\mathrm{e}{-4}$.
We trained the model for $30$ epochs until it converged and decreased the learning rate by $0.2$ times after $15$ epochs.
To mitigate the fitting of the model against its training data, we set the weight decay factor to $1\mathrm{e}{-5}$.

\noindent \textbf{Evaluation.}
To evaluate the QA performance, we used exact matches EM@1 and EM@10 as the evaluation metric, where EM@$K$ is the percentage of predictions in which the top $K$ predicted answers exactly match any one of the ground-truth answers.
We also included sentence evaluation metrics frequently used for image captioning models because some of the questions had multiple possible answer expressions, as discussed in Sec.~\ref{sec:dataset_stat}. We added the BLEU~\cite{bleu}, ROUGE-L~\cite{rouge}, METEOR~\cite{meteor}, CIDEr~\cite{cider}, and SPICE~\cite{spice} metrics to evaluate robust answer matching.

\noindent \textbf{Baselines.}
To validate our 3D-QA model (ScanQA), we prepared several baselines. Empirical experiments were conducted using the following methods.

\noindent \textit{\textbf{RandomImage+2D-QA}}
First, we prepared several 2D-QA models as baselines to demonstrate how our 3D-QA models outperformed 2D-based VQA models for the 3D-QA task. 
We used a pretrained {MCAN} model ~\cite{Yu_2019_CVPR} as the 2D-QA model.
MCAN is a transformer network~\cite{NIPS2017_3f5ee243} that uses a cross-attention mechanism to represent the relationship between question words and objects in an image. 
The proposed method uses some of the modules used in MCAN, such as transformer encoder and decoder layers, to create 3D and language features. By comparing these two, we can confirm the importance of creating a model specialized for 3D-QA.
Because 2D-QA models cannot be directly applied to a 3D environment, we randomly sampled three images from the video captured to build the ScanNet dataset. We used a bottom-up top-down attention model~\cite{Anderson_2018_CVPR} to extract the appearance features of the objects. We applied pretrained 2D-QA models to these images and computed the answer scores for each image. Finally, we selected the most probable answer according to the averaged answer scores of these images. 
We experimented with 2D-QA using three images captured in the environment per question.

\noindent \textit{\textbf{OracleImage+2D-QA}}
To investigate the upper bound on the performance of 2D-QA for questions in 3D space, we used images around a target object associated with a question--answer pair.
We set the camera's position based on the coordinates of the bounding box of the correct object and captured images from the direction and distance at which the bounding box was most visible. Because the object may not be visible depending on the camera's position, we used three images per question.
We applied 2D-QA models to these images similar to RandomImage+2D-QA.
Note that it is difficult to obtain such images in actual QA scenarios.
By examining the performance of this method, we can determine the difficulty of solving the 3D-QA task using 2D-QA models.

\noindent \textit{\textbf{VoteNet+MCAN}}
VoteNet~\cite{Qi_2019_ICCV} is a 3D object detection method that locates and recognizes objects in a 3D scene.
This method detects objects in a 3D space, extracts their features, and uses them in a standard VQA model (MCAN).
Unlike our method, this method does not consider the target object or its location in the 3D space.

\begin{figure*}
 \centering
 \includegraphics[keepaspectratio, scale=0.28]{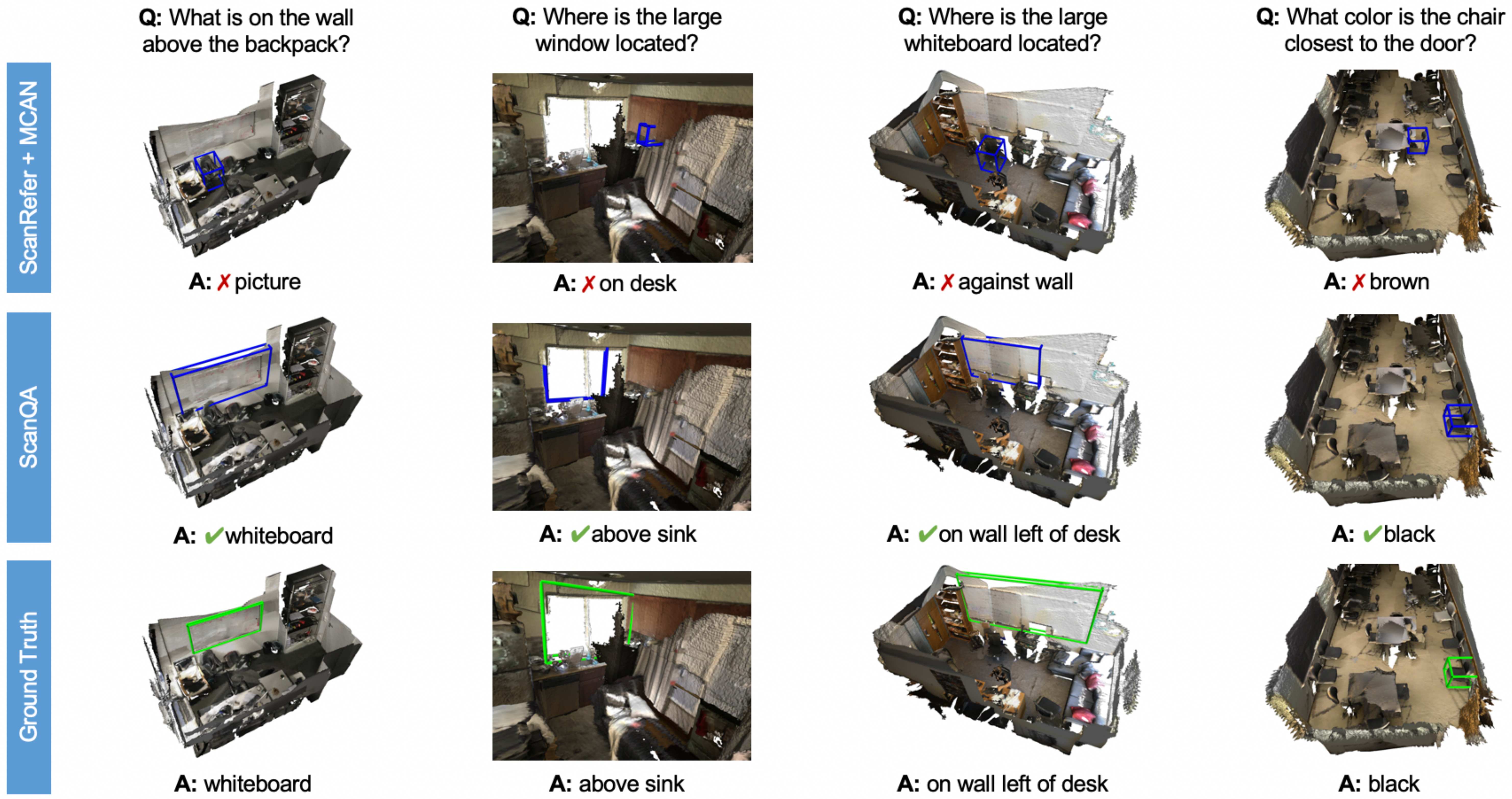} 
   \caption{\textbf{Qualitative results.} 
Predicted answers are described below each figure.
Predicted boxes are marked blue and the ground truth is marked green.
    We show examples in which ScanQA produced good object localization when predicting correct answers, whereas ScanRefer+MCAN (pipeline) could not.
  }
 \label{fig:qualitative}
 \vspace{-4mm}
\end{figure*}

\noindent \textit{\textbf{ScanRefer+MCAN (pipeline)}}
ScanRefer\cite{chen2020scanrefer} is a 3D object localization method for localizing a given linguistic description to a corresponding target object in a 3D space. ScanRefer internally uses VoteNet to detect objects in a room and estimates the object corresponding to the linguistic description from among the candidate objects.
Note that ScanRefer cannot be used directly for QA.
Thus, we used a pretrained ScanRefer model 
to identify the object corresponding to the question and then applied 2D-QA (MCAN) to the image surrounding the object localized by ScanRefer.
Note that ScanRefer and MCAN were run separately, and end-to-end learning was not possible.

\noindent \textit{\textbf{ScanRefer+MCAN (end-to-end)}}
This method is more sophisticated and closer to the proposed method.
Although ScanRefer+MCAN (pipeline) conducts object localization and QA separately, this method simultaneously learns localization and QA modules.
Specifically, the input to the method is the object proposal feature of ScanRefer for VQA models; subsequently, the model 
predicts answers based on object box features and question content.
Unlike the ScanQA model, this uses the output of VoteNet separately for object localization and QA modules (although the information for both tasks is mutually useful) and does not learn both modules in common.

\subsection{Quantitative Analysis}
The performance of 3D-QA on the ScanQA dataset and image caption metrics are presented in Table~\ref{table:performance3D}.
The best results in each column are shown in bold.
We compared our ScanQA model with competitive baselines VoteNet+MCAN, ScanRefer+MCAN (pipeline), and ScanRefer+MCAN (end-to-end).
These baselines share some of the components of the proposed method and aided us 
in understanding useful components for 3D-QA.
The results indicated that our ScanQA method significantly outperformed all baselines across all data splits over all evaluation metrics. In particular, 
ScanQA significantly outperformed ScanRefer+MCAN (end-to-end), which learns QA and object localization separately across all data splits over all evaluation metrics, indicating that the ScanQA model succeeded in synergistic learning by sharing object localization and QA modules.
ScanQA outperformed VoteNet+MCAN by a large margin. This result suggested that using object localization in a 3D space and predicting object categories related to questions are important for 3D-QA. We will clarify this point in the section on the ablation study.
Interestingly, VoteNet+MCAN, ScanRefer+MCAN (end-to-end), and ScanQA significantly outperformed ScanRefer+MCAN (pipeline), 
which detects target objects related to a question using a pretrained ScanRefer and then applies 2D-QA to the surrounding images of a target object. The results indicated that end-to-end training with 3D and language information is suitable for solving 3D-QA model problems.
In addition, we observed that our 3D-QA model, ScanQA, is superior to a 2D-QA model, RandomImage+MCAN, which uses an effective pretrained model.
We also observed that the 2D-QA baseline with oracle object identification of OracleImage+MCAN performed better or more competitively than the ScanQA model. Although this suggests that accurate object identification for questions indeed boosts 3D-QA results, this is an oracle setting for real-world applications. 
We finally evaluated the human performance on the sampled questions in the test set with objects using MTurk. The exact matching score (EM@1) is 51.6 for the best-performing MTurk worker.

\subsection{Ablation Studies}
We conducted ablation studies concerning the design of the ScanQA model.
Table~\ref{table:component_ablation} lists the effects of each major component of the proposed method. 
The effect of different input data is shown in Table~\ref{table:feature_ablation}.

\noindent \textbf{Does object classification help?} 
We demonstrated the effectiveness of object classification by conducting an experiment using ScanQA combined with and without the object classification module. We compared our method trained with the answer and object classification modules (ANS+OBJ) with a model trained with only the answer module (ANS), and we also compared a model trained with full modules (ANS+OBJ+LOC) with one trained with the answer and object localization modules (ANS+LOC). 
The results in Table~\ref{table:component_ablation} show that the models with the OBJ outperformed the other models. 
This suggests that predicting the category of a target object is effective for 3D-QA.

\noindent \textbf{Does object localization help?} 
We demonstrated the effectiveness of object localization by conducting an experiment with ScanQA and without an object localization module. We compared our method trained with the answer and object localization modules (ANS+LOC) with a model trained with only the answer module (ANS), and we also compared a model trained with full modules (ANS+OBJ+LOC) with one trained with the answer and object classification modules (ANS+OBJ). 
The results in Table~\ref{table:component_ablation} show that the model with LOC consistently outperformed the other models. 
This suggests that localizing target objects is also important for improving 3D-QA performance.

\noindent \textbf{Do colors help for 3D-QA?}
According to the ScanRefer study, models that use color information have better object localization performance than models that use only geometry~\cite{chen2020scanrefer}.
Motivated by this finding, we evaluated several models using different features.
We compared our method trained with geometry (xyz) and multiview image features (xyz+multiview) with a model trained with only geometry (xyz) and one trained with RGB values
(xyz+rgb).
To validate the object localization performance, we used accuracy (Acc@0.25), in which the positive predictions have a higher intersection over union with the ground truths than the threshold of 0.25 used in~\cite{chen2020scanrefer}.
As Table~\ref{table:feature_ablation} shows,  RGB values were not effective for both object localization and QA.
The multiview image features were slightly effective for object localization but not on QA.
This is because it is more difficult for the ScanQA dataset to associate language and object information than the ScanRefer dataset because multiple objects may apply to a single question.
Fortunately, ScanQA trained with geometry, preprocessed multiview image features, and normals 
which is demonstrated in Table~\ref{table:performance3D}
outperformed the other models, but the effect was limited.
This result suggested that subsequent studies should consider a more balanced selection of features in terms of computational cost and performance.

\subsection{Qualitative Analysis}
Finally, we demonstrated the excellent performance of our model by visualizing qualitative examples of ScanQA, ScanRefer+MCAN (pipeline), and ground truth.
Fig.~\ref{fig:qualitative} shows the representative QA results of a baseline method and ScanQA. 
The results suggested that answering questions requires object localization related to the answer according to the question content and point cloud matching. 
For example, the leftmost case shows that a whiteboard located above a backpack could not be answered by ScanRefer+MCAN (pipeline), which localized the backpack in error.
This is because the QA and localization modules were separated in ScanRefer+MCAN (pipeline). In contrast, our model successfully localized target objects related to answers and predicted correct answers by simultaneously learning QA and object localization.

%% file: sec/conclusion.tex
\section{Conclusion}
Spatial understanding using language expression is a core technology for models deployed in the real world and interacting with humans.
We introduce a novel task of 3D question answering (3D-QA), in which models observe an entire 3D scan and answer a question about the 3D scene in addition to the object localization.
Based on the ScanRefer dataset, we created a new ScanQA dataset which consists of 41,363 questions and 32,337 unique answers from 800 scenes derived from the ScanNet scenes.
We propose a 3D-QA baseline model of ScanQA. We confirm that the ScanQA baseline performs better than the counterpart 2D-based VQA baselines in most of the evaluation measures, including the exact match and image captioning metrics.

\section*{Acknowledgements}
This work was supported by NEDO JPNP20006, JSPS KAKENHI 21H03516 and 18KK0284, and JST PRESTO JPMJPR20C2.

%% file: sec/X_supplementary.tex
\clearpage
\appendix

\setcounter{page}{1}

\twocolumn[
\centering
\Large
\textbf{Supplementary Material} \\
\vspace{1.0em}
] 
\appendix

\if[]
\begin{figure*}[t]
 \centering
 \includegraphics[keepaspectratio, scale=0.72]{fig/baseline_overview.pdf} 
 \caption{Illustration of the baselines and our model for answering 3D environments}
 \label{fig:baseline_ovierview}
\end{figure*}
\fi

In this supplementary material, 
we first provide additional experimental details in Section~\ref{sec:supp_exp_setup}. 
Next, we provide additional experimental quantitative (Section ~\ref{sec:supp_quantitative}) and qualitative results (Section~\ref{sec:supp_qualitative}). Finally, we have attached a snapshot of the annotation website (Section~\ref{sec:supp_mturk}).

\section{Additional Experimental Details}~\label{sec:supp_exp_setup}
We conducted additional experiments to demonstrate the effectiveness of the proposed method compared with various other methods. In this section, we introduce other baselines and an additional ScanQA model that considers multiple objects related to a question.

\subsection{Additional Baseline Models}
\begin{figure}[t]
 \centering
 \includegraphics[keepaspectratio, scale=0.46]{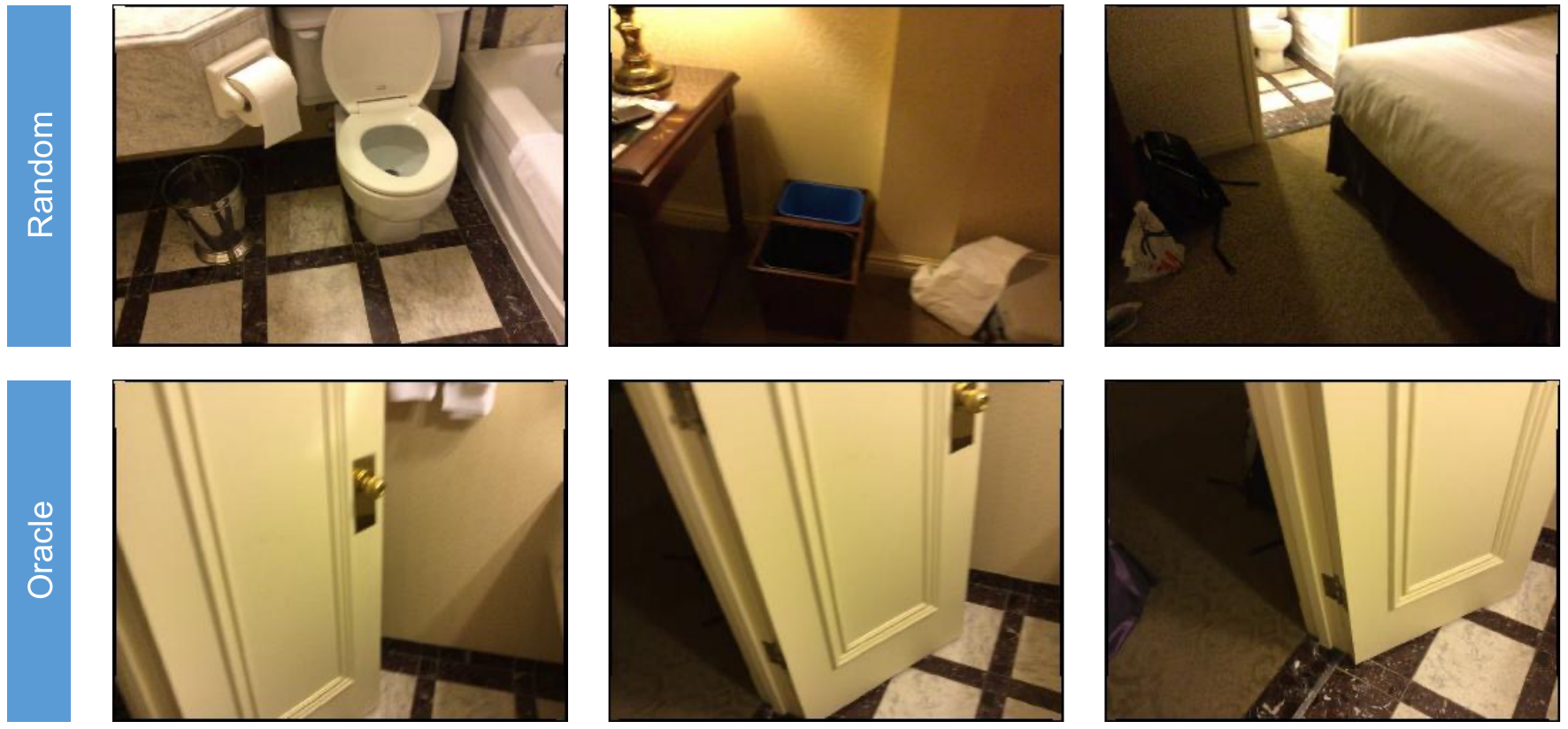} 
 \vspace{-0.6cm}
 \caption{
Example of real images about the question ``What color is the bathroom door?'' The upper panel is RandomImage, and the lower panel is OracleImage. 
 }
 \label{fig:real_image}
 \vspace{-0.2cm}
\end{figure}

\begin{figure}[t]
 \centering
 \includegraphics[keepaspectratio, scale=0.46]{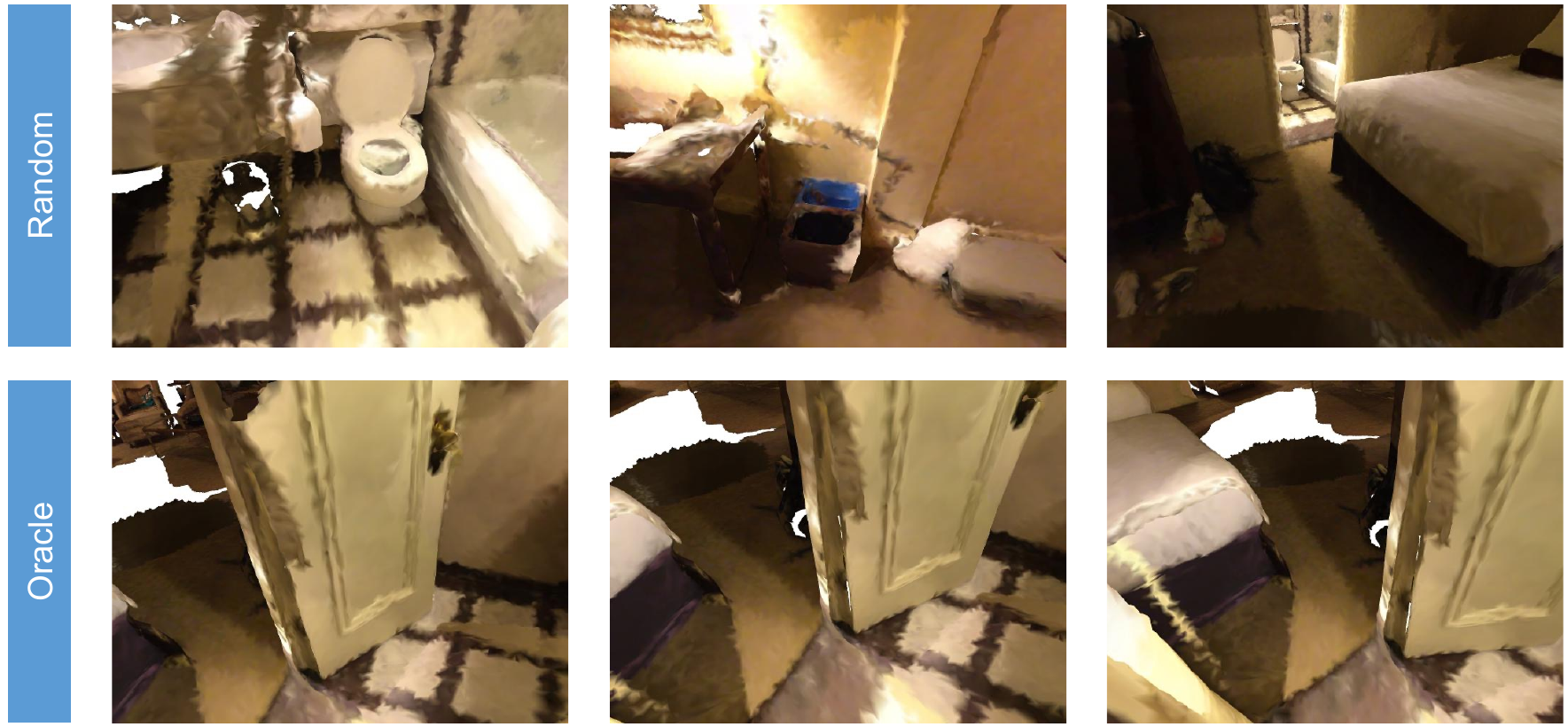} 
 s\vspace{-0.6cm}
  \caption{
Example of mesh images about a question ``What color is the bathroom door?'' The upper panel is RandomImage, and the lower panel is OracleImage. 
 }
  \label{fig:mesh_image}
 \vspace{-0.2cm}
\end{figure}

\begin{figure}[t]
 \centering
 \includegraphics[keepaspectratio, scale=0.42]{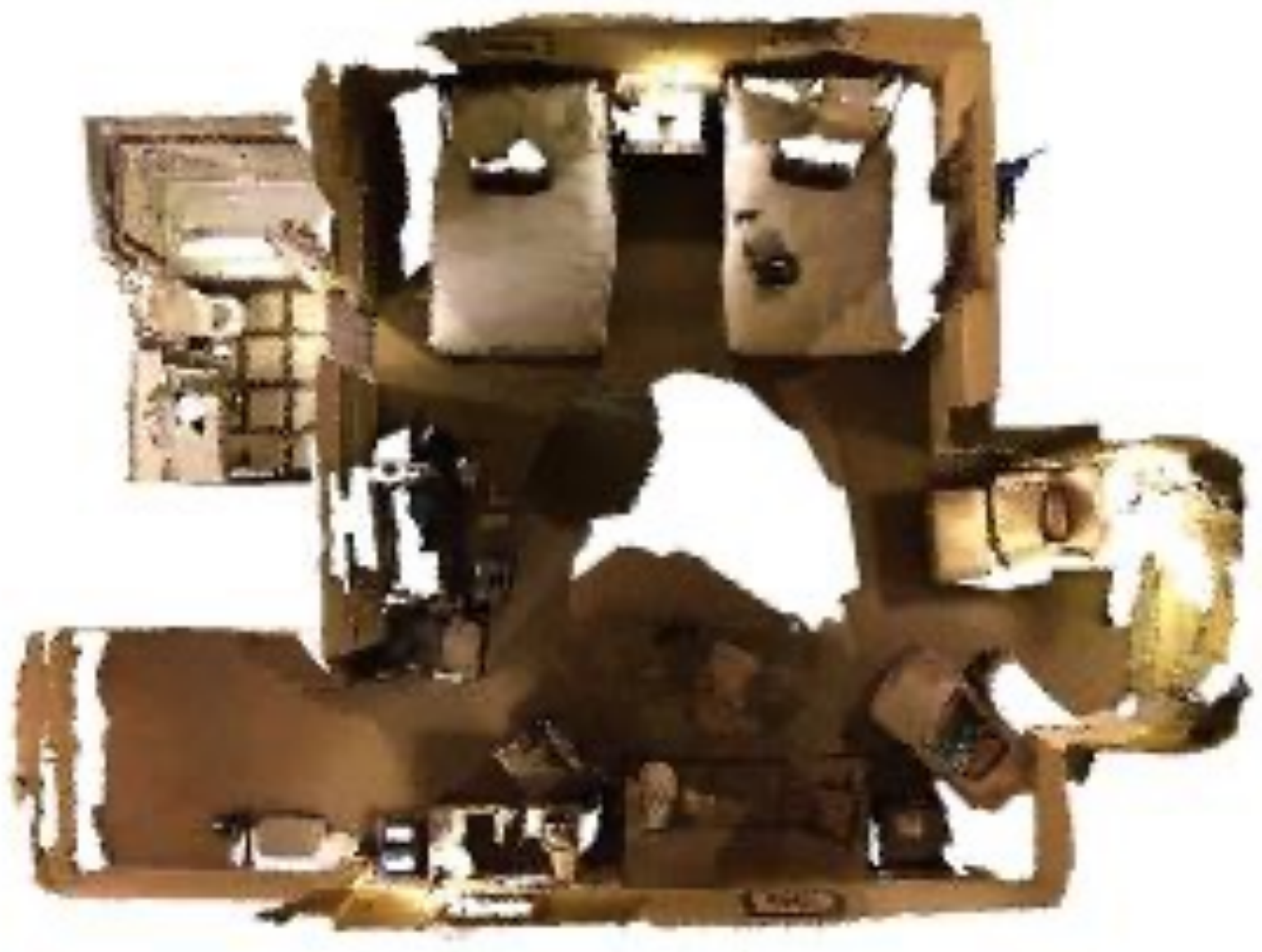} 
 \vspace{-0.2cm} 
 \caption{
Example of a TopDownImage about the question ``What color is the bathroom door?'' 
 }
 \label{fig:top_down_image}
 \vspace{-0.2cm} 
\end{figure}

\noindent \textbf{Additional 2D-QA Image.}
We prepared RandomImage (or OracleImage) + 2D-QA as a 2D-QA baseline method that uses three images captured in the environment per question. 
Fig.~\ref{fig:real_image} shows the images randomly sampled from the video to build the ScanNet dataset. In addition to using such real images from the environments, we also used images of the mesh data of ScanNet captured at positions and angles similar to real images as in Fig.~\ref{fig:mesh_image}.
We refer to models using real images ``real" and ones using mesh images ``mesh." We also used mesh images captured from a top-down view (referred to as TopDownImage) to view the entire room with a single image (Fig.~\ref{fig:top_down_image}).

\noindent \textbf{Additional 2D-QA Model.}
In addition to MCAN~\cite{Yu_2019_CVPR}, we evaluated 2D-QA using the BERT-based model Oscar~\cite{li2020oscar}
trained with many image-text pairs and has demonstrated high performance in various tasks, such as VQA, image retrieval, image captioning, and natural language visual reasoning.
Although MCAN and Oscar use effective pretrained models, unlike our method, we can emphasize the effectiveness of the ScanQA model by comparing it to these models.

\subsection{Additional ScanQA Model}
The ScanQA model introduced in Section~\ref{sec:proposed_method} predicts the object confidence (box confidences) and object classification for a single object.
However, a given question is occasionally associated with more than one object.
Thus, we extended the ScanQA model to perform object localization and labeling of multiple objects. Hence, we computed the final scores for all object proposals (or object labels) normalized by a sigmoid function and used BCE loss to train both the object localization and object classification modules. The proposed method used in this study was ScanQA (single), and the model that considers multiple objects was called ScanQA (multiple).  Hereafter, unless otherwise specified, ScanQA is referred to as ScanQA (single).

\section{Additional Quantitative Experiments}~\label{sec:supp_quantitative}
\vspace{-0.6cm}
\subsection{Object Localization Results}
Before introducing the additional QA results, we demonstrated the object localization performance using ScanRefer+MCAN (pipeline), 
ScanRefer+MCAN (end-to-end), and ScanQA. 
We used the Acc@$K$, where the positive predictions had a higher IoU with the ground truths than the threshold $K$ (set to 0.25 and 0.5)~\cite{chen2020scanrefer}.
As shown in Table~\ref{table:feature_ablation}, we refer to each accuracy as Acc@0.25 and Acc@0.5 in our experiments.
The results in Table~\ref{table:object_localization_performance}
show that the shared and end-to-end learning of QA, object localization, and object classification modules effectively predicted the target object for a given question.

\input{supp_result/object_localization}

\subsection{Question Answering Results}
Table~\ref{table:additional_overall_performance} shows the performance of additional baselines and proposed models on the ScanQA dataset. The best results in each column are indicated in bold. The results show that our ScanQA (single or multiple) models outperformed the baselines RandomImage + MCAN (mesh), RandomImage + Oscar (mesh), TopDownImage + MCAN, and TopDownImage + Oscar across all evaluation metrics on all splits. While RandomImage + Oscar (real) outperformed ours on EM@1 on the test without objects split owing to its effective pretrained model, the ScanQA models outperformed RandomImage + Oscar (real) on other evaluation metrics. Regarding the difference using real and mesh images, the performance tended to be better when using real images.
For example, RandomImage + MCAN (real) outperformed
RandomImage + MCAN (mesh) and RandomImage + Oscar (real) outperformed RandomImage + Oscar (mesh) in almost all evaluation measures on all splits. 
We observed no consistency in the advantage regarding the performance difference between ScanQA (single) and ScanQA (multiple).

\subsection{Ablation Studies on ScanQA (multiple)}
In this section, we describe ablation studies conducted on the ScanQA (multiple) model that predicted confidences and labels for multiple objects when performing QA.
The effect of each major component of the proposed method is shown 
in Table~\ref{table:scanqa_multi_component_ablation}, 
and the effect of different input data is shown in Table~\ref{table:scanqa_multi_feature_ablation}.
The results in Table~\ref{table:scanqa_multi_component_ablation} suggest that using the object localization module (LOC) or the object classification module (OBJ) was effective for improving QA performance.
Table~\ref{table:scanqa_multi_feature_ablation} shows the object localization performance Top10-Acc@0.25 and QA performance EM@10 on ScanQA (multiple) using different input features, where
Top10-Acc@0.25 is the accuracy at which the positive predictions had higher IoU with the ground truth than 0.25 (we compare the top 10 object boxes with the highest object localization scores with the ground true boxes and consider positive predictions for the box with the highest IoU.)
We observed that RGB values were effective for QA, and 
the multiview image features were effective for both object localization and QA. 
We assumed that by predicting multiple objects, ScanQA (multiple) 
could utilize those various features.

\input{supp_result/multi_feature_ablation}

\subsection{Performance by Parameter}
We also evaluated the performance of ScanQA with different parameters, the number of layers $L$, and the hidden size $d$ in Tables ~\ref{table:num_layer} and ~\ref{table:hidden_size}, respectively.
The results suggest that the number of layers $L=1$ and  hidden size $d=128$ were suitable for the test splits.

\input{supp_result/overall_result}

\input{supp_result/multi_module_ablation}

\input{supp_result/num_layer}
\input{supp_result/hidden_size}
\input{supp_result/question_type}

\begin{figure*}[t]
 \centering
 \includegraphics[keepaspectratio, scale=0.28]{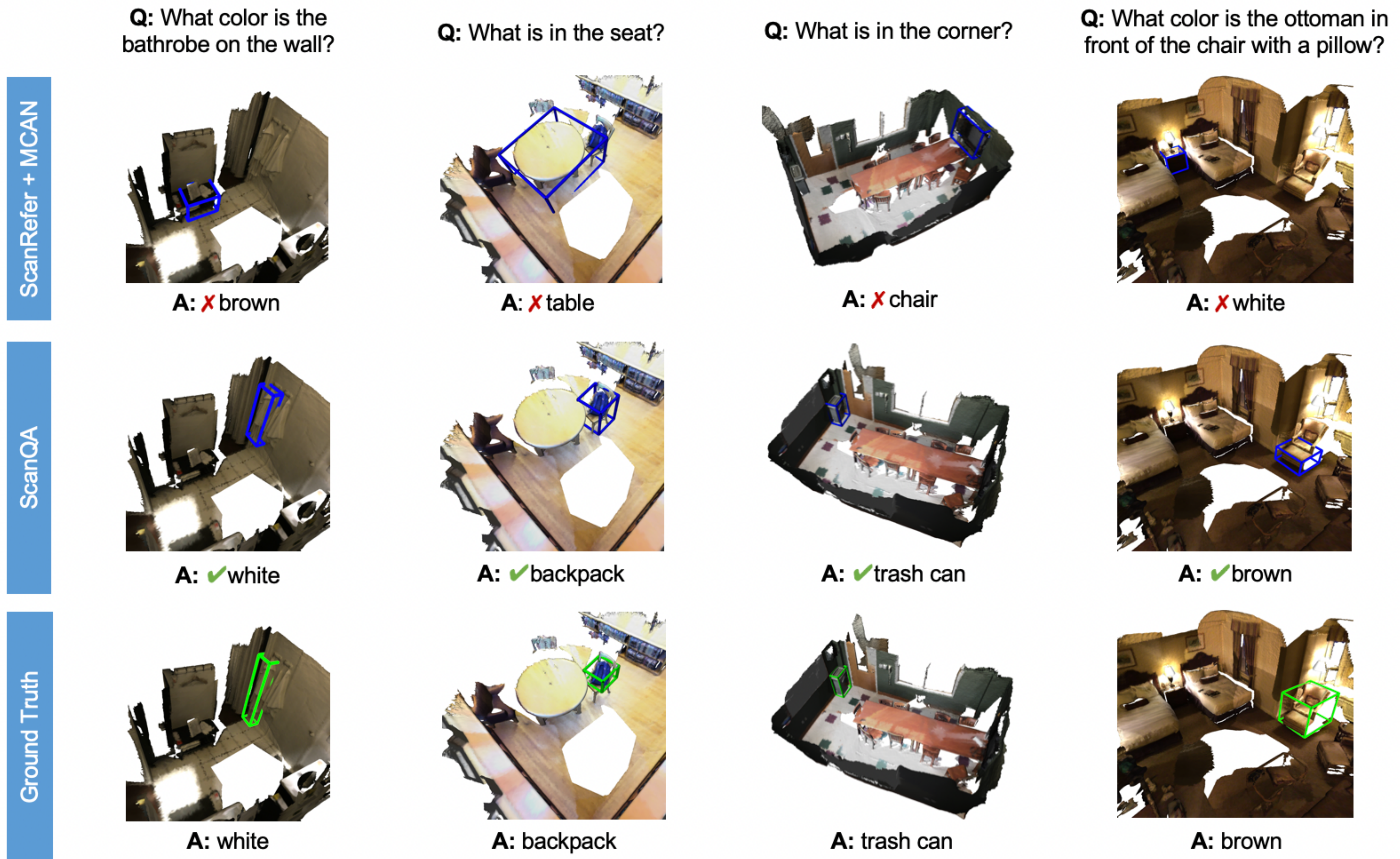} 
 \caption{
    Qualitative analysis comparing ScanQA and ScanRefer + MCAN (end-to-end).
 }
 \label{fig:additional_qualitative_experiments}
\end{figure*}

\section{Additional Qualitative Analysis}~\label{sec:supp_qualitative}
We conducted a qualitative evaluation by comparing our ScanQA model to 
the ScanRefer + MCAN (end-to-end)
in addition to the evaluation on ScanRefer + MCAN (pipeline) (Fig.~\ref{fig:qualitative}).
Fig.~\ref{fig:additional_qualitative_experiments} shows the results of the object localization and QA correctness with the visualization of the scene and bounding boxes using ScanRefer + MCAN (end-to-end), ScanQA, and ground truth.
The results also indicated that QA correctness and localization were closely related.
The leftmost case shows that models were to answer the color of the bathrobe.
However, the baseline model, ScanRefer + MCAN (end-to-end), localized a different object ``table'' and answered its color. 
The baseline model used the word ``wall'' to determine an object near the wall and localized it.
However, our model could localize the correct object, ``bathrobe,'' on the wall.
In the second case from the left, the question was about the object placed on the chair, but the baseline model provided the wrong answer ``table''
because the word ``seat'' is frequently associated with the table.
In contrast, our model understood the meaning ``in the seat'' and correctly selected a backpack in the seat.
In the second case from the right,
the baseline model incorrectly localized a TV close to the wall and answered ``chair'' close to the TV.
Our model correctly recognized the meaning ``corner," localized it, and indicated the correct object ``trash can'' at the corner.
In the final case, there were multiple ottomans in the scene, and the models were to correctly understand the positional relationship between the ottoman, chair, and pillow.  
The baseline model localized the lamp and incorrectly answered its color ``white.''
In contrast, our model correctly understood the positional relationship between the ottoman, chair, and pillow, localizing the ottoman in front of the chair with a pillow.

\section{MTurk Annotation Details}
\label{sec:supp_mturk}
\begin{figure*}[t]
 \centering
 \includegraphics[keepaspectratio, scale=0.5]{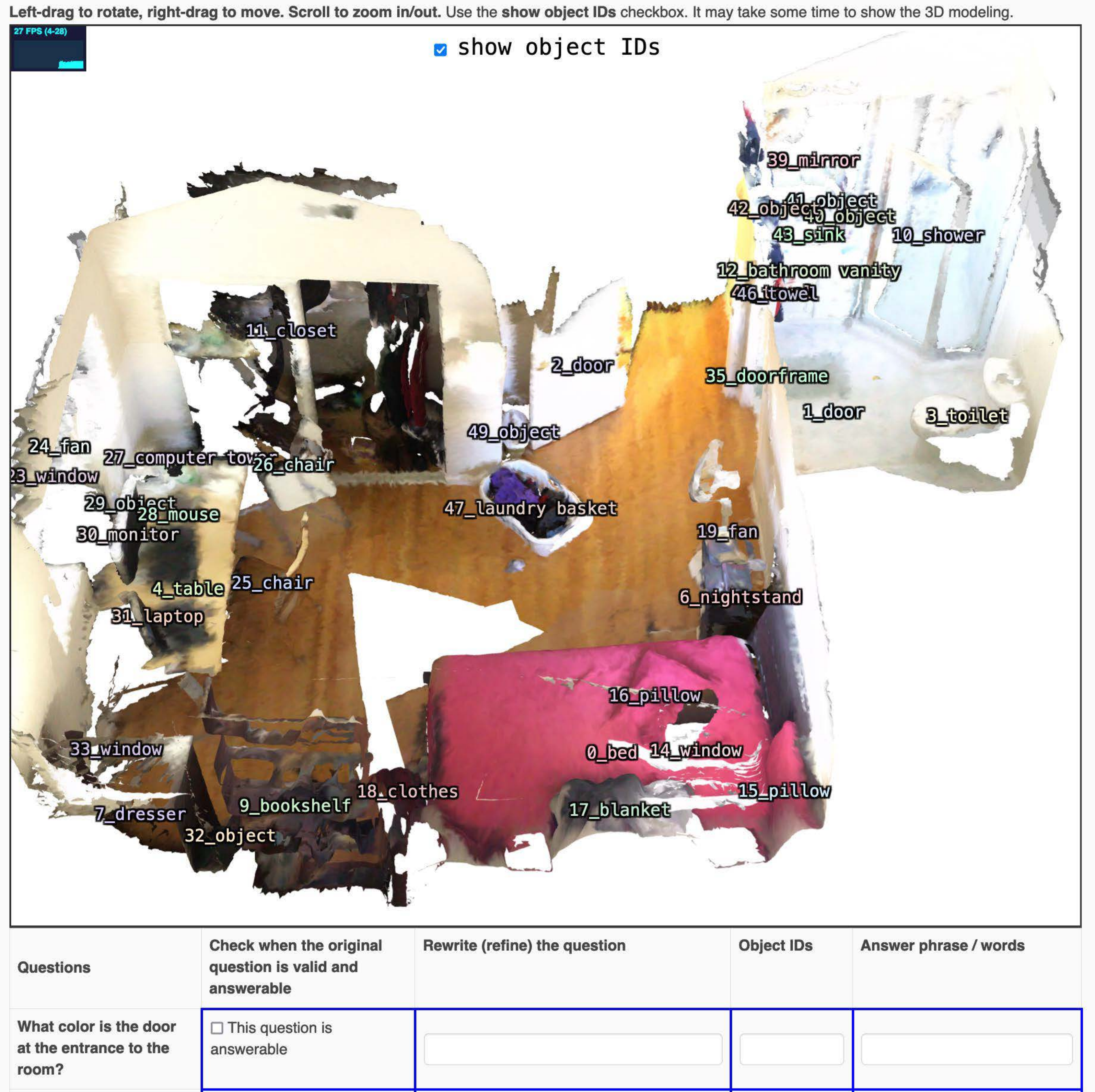}
 \caption{
Example of 3D modeling viewer and QA form on the MTurk website.
 }
 \label{fig:mturk_snapshot}
\end{figure*}

We developed a visualizer website for 3D modeling. MTurk workers can interactively rotate and zoom in 3D modeling.
Fig.~\ref{fig:mturk_snapshot} presents the snapshot of the MTurk website for the editing and answer collection phrase of the QA annotation.

We filtered the auto-generated questions as follows. We first applied rule-based filtering for removing potential underspecified or noisy expressions such as ``this’’ and ``image'' and direction words, namely ``north,’’ ``west,’’ ``south,’’ and ``east.’’
We also filtered odd question types, such as ``What is the name of...’’
Furthermore, we filtered meaningless questions in 3D scenes using MTurk.
We asked three workers to evaluate each question and filtered those questions that were evaluated as ``valid'' in the scene by at least two of the three workers.
Finally, we asked workers to rewrite those questions that, according to them, were underspecified before they filled out the answers.

We consider that our dataset covers a broad range of questions with distinct meanings as they are constructed through human (re)annotation; they reflect various questions that humans may ask. The number of unique answers is also high corresponding to the number of unique questions.
We noticed that the question length in terms of tokens had a fat tail distribution.
Interestingly, there are both very short and long questions in the dataset such as
``How many chairs?'' and
``What color is the chair that is located to the right of another brown chair with a red bag on it?''

%% file: supp_result/object_localization.tex
\begin{table}[t]
\begin{center}
	\footnotesize\begin{tabular}{lcc}
        \toprule
Model                & Acc@0.25 &  Acc@0.5 \\
\midrule
\textbf{Valid} \\
ScanRefer+MCAN (pipeline) 	 &  12.88 & 9.13 \\
ScanRefer+MCAN (end-to-end) 	 &  23.53 & 11.76 \\
ScanQA 	 &  \textbf{24.96} & \textbf{15.42} \\
\midrule
\textbf{Test w/ objects} \\
ScanRefer+MCAN (pipeline) 	 &  12.94 & 8.02 \\
ScanRefer+MCAN (end-to-end) 	 &  21.97 & 10.41 \\
ScanQA 	 &  \textbf{25.44} & \textbf{15.03} \\
\bottomrule
	\end{tabular}
    \caption{Object localization performance on the ScanQA dataset}
    \label{table:object_localization_performance}
\end{center}
\vspace{-4mm}
\end{table}

%% file: supp_result/multi_feature_ablation.tex
\begin{table}[t]
\begin{center}
	\footnotesize\begin{tabular}{lcc}
        \toprule
Model                & Top10-Acc@0.25 &  EM@10 \\
\midrule
\textbf{Valid} \\
ScanQA (xyz)  	 &  67.83 & 49.58 \\
ScanQA (xyz+rgb)  	 &  66.72 & 50.22 \\
ScanQA (xyz+rgb+normal)   	 &  68.13 & 49.45 \\
ScanQA (xyz+multiview)  	 &  67.85 & 49.86 \\
ScanQA (xyz+multiview+normal) 	 &  \textbf{70.82} & \textbf{51.23} \\
\midrule
\textbf{Test w/ objects} \\
ScanQA (xyz)  	 &  68.23 & 55.18 \\
ScanQA (xyz+rgb)  	 &  68.87 & \textbf{56.23} \\
ScanQA (xyz+rgb+normal)   	 &  69.65 & 55.25 \\
ScanQA (xyz+multiview)  	 &  70.76 & 55.65 \\
ScanQA (xyz+multiview+normal) 	 &  \textbf{71.58} & 55.99 \\
\bottomrule
	\end{tabular}
    \caption{
        Feature ablation results on ScanQA (multiple)
    }
    \label{table:scanqa_multi_feature_ablation}
\end{center}
\vspace{-4mm}
\end{table}

%% file: supp_result/overall_result.tex
\begin{table*}[t]
\begin{center}
\scriptsize
\begin{tabular}{lcccccccccc}
        \toprule
Model                & EM@1 & EM@10 & BLEU-1 & BLEU-2 & BLEU-3 & BLEU-4 & ROUGE & METEOR & CIDEr & SPICE \\
\midrule
\textbf{Valid} \\
RandomImage+MCAN (real) & 19.19 & 48.15 & 23.71 & 15.41 & 11.81 & 0.00 & 28.90 & 10.92 & 53.83 & 9.35 \\
RandomImage+MCAN (mesh) & 18.59 & 47.81 & 22.12 & 14.49 & 11.72 & 7.69 & 27.61 & 10.32 & 50.93 & 8.37 \\
RandomImage+Oscar (real) & 19.38 & 46.37 & 22.91 & 14.52 & 11.20 & 0.56 & 28.70 & 10.66 & 52.86 & 8.91 \\
RandomImage+Oscar (mesh) & 17.97 & 43.55 & 20.67 & 12.01 & 11.51 & 0.57 & 26.51 & 9.82 & 47.82 & 7.75 \\
TopDownImage+MCAN & 12.71 & 41.50 & 14.82 & 8.21 & 16.54 & 0.74 & 19.33 & 7.57 & 33.14 & 5.72 \\
TopDownImage+Oscar & 17.20 & 43.81 & 19.75 & 11.21 & 15.31 & 0.71 & 25.39 & 9.43 & 45.21 & 7.32 \\
VoteNet+MCAN 	 &  17.33 & 45.54 & 28.09 & 16.72 & 10.75 & 6.24 & 29.84 & 11.41 & 54.68 & 10.65 \\
ScanRefer+MCAN (pipeline, real) & 14.37 & 44.12 & 17.02 & 10.17 & 15.77 & 0.72 & 22.02 & 8.45 & 38.73 & 6.66 \\
ScanRefer+MCAN (pipeline, mesh) & 14.57 & 43.27 & 16.71 & 9.71 & 13.62 & 0.64 & 21.82 & 8.32 & 38.35 & 6.49 \\
ScanRefer+MCAN (e2e) 	 &  18.59 & 46.76 & 26.93 & 16.59 & 11.59 & 7.87 & 30.03 & 11.52 & 55.41 & 11.28 \\
ScanQA (single) &  {20.28} & {50.01} & {29.47} & {19.84} & {14.65} & {9.55} & {32.37} & {12.60} & {61.66} & {11.86} \\
ScanQA (multiple) 	&  \textbf{21.05} & \textbf{51.23} & \textbf{30.24} & \textbf{20.40} & \textbf{15.11} & \textbf{10.08} & \textbf{33.33} & \textbf{13.14} & \textbf{64.86} & \textbf{13.43} \\

\midrule
OracleImage+MCAN (real) & 22.59 & 49.43 & 26.58 & 18.32 & 15.37 & 8.50 & 33.23 & 12.45 & 63.44 & 12.56 \\
OracleImage+MCAN (mesh) & 20.66 & 48.04 & 24.35 & 17.00 & 14.23 & 0.00 & 30.34 & 11.23 & 57.01 & 10.15 \\
OracleImage+Oscar (real) & 21.39 & 45.05 & 24.29 & 14.19 & 14.21 & 0.67 & 31.24 & 11.33 & 58.23 & 10.51 \\
OracleImage+Oscar (mesh) & 22.27 & 46.59 & 23.01 & 13.96 & 14.23 & 0.00 & 31.37 & 11.53 & 57.98 & 11.11 \\
\midrule
\textbf{Test w/ objects} \\
RandomImage+MCAN (real) & 22.31 & 53.11 & 26.66 & 18.49 & 16.16 & 14.26 & 31.27 & 12.13 & 60.37 & 9.05 \\
RandomImage+MCAN (mesh) & 21.74 & 52.41 & 24.86 & 17.49 & 15.33 & 15.19 & 30.00 & 11.55 & 57.55 & 8.73 \\
RandomImage+Oscar (real) & 22.65 & 52.35 & 24.74 & 14.42 & 9.85 & 0.00 & 30.81 & 11.59 & 57.72 & 8.52 \\
RandomImage+Oscar (mesh) &20.92 & 49.22 & 23.18 & 12.26 & 9.06 & 0.48 & 28.95 & 10.86 & 53.11 & 7.12 \\
TopDownImage+MCAN     & 15.76 & 47.15 & 16.52 & 8.59 & 0.00 & 0.00 & 21.49 & 8.37 & 38.55 & 5.34 \\
TopDownImage+Oscar    & 20.76 & 49.26 & 22.19 & 11.02 & 9.30 & 0.49 & 28.25 & 10.53 & 51.82 & 6.83 \\
VoteNet+MCAN 	    &  19.71 & 50.76 & 29.46 & 17.23 & 10.33 & 6.08 & 30.97 & 12.07 & 58.23 & 10.44 \\
ScanRefer+MCAN (pipeline, real) & 17.52 & 49.92 & 19.17 & 10.66 & 0.00 & 0.00 & 24.40 & 9.38 & 44.25 & 6.24 \\
ScanRefer+MCAN (pipeline, mesh) & 17.44 & 48.83 & 18.45 & 9.49 & 0.00 & 0.00 & 23.90 & 9.11 & 42.97 & 5.93 \\
ScanRefer+MCAN (e2e) 	 &  20.56 & 52.35 & 27.85 & 17.27 & 11.88 & 7.46 & 30.68 & 11.97 & 57.36 & 10.58 \\
ScanQ (single) 	&  \textbf{23.45} & \textbf{56.51} & \textbf{31.56} & \textbf{21.39} & \textbf{15.87} & \textbf{12.04} &
\textbf{34.34} & {13.55} & \textbf{67.29} & {11.99} \\
ScanQA (multiple)  &  23.05 & 55.99 & 31.40 & 21.18 & 15.82 & 11.70 & 34.05 & \textbf{13.60} & 66.76 & \textbf{12.30} \\
\midrule
OracleImage+MCAN (real) & 25.34 & 55.93 & 28.70 & 20.11 & 16.78 & 12.89 & 34.59 & 13.42 & 67.24 & 11.93 \\
OracleImage+MCAN (mesh) & 23.35 & 53.05 & 25.90 & 17.15 & 13.36 & 10.94 & 31.66 & 12.08 & 60.64 & 9.01 \\
OracleImage+Oscar (real) & 25.30 & 50.12 & 26.38 & 14.10 & 8.70 & 0.47 & 33.72 & 12.47 & 63.31 & 9.48 \\
OracleImage+Oscar (mesh) & 25.52 & 52.39 & 25.17 & 14.13 & 10.94 & 0.00 & 33.51 & 12.68 & 63.13 & 10.52 \\
\midrule
\textbf{Test w/o objects} \\
RandomImage+MCAN (real) & 20.82 & 51.23 & 26.29 & 17.90 & 14.27 & 9.66 & 29.23 & 11.54 & 55.64 & 8.87 \\
RandomImage+MCAN (mesh) & 20.34 & 51.20 & 24.72 & 16.93 & 14.04 & 7.55 & 28.10 & 10.95 & 53.41 & 8.61 \\
RandomImage+Oscar (real) & \textbf{21.58} & 49.85 & 24.86 & 16.16 & 13.29 & 0.00 & 28.99 & 10.99 & 54.62 & 8.62 \\
RandomImage+Oscar (mesh) & 19.91 & 48.74 & 23.02 & 13.24 & 12.91 & 0.64 & 26.94 & 10.17 & 49.83 & 7.45 \\
TopDownImage+MCAN     & 14.39 & 45.94 & 15.70 & 8.86 & 12.56 & 0.62 & 19.26 & 7.71 & 34.59 & 5.22 \\
TopDownImage+Oscar    & 19.13 & 48.06 & 21.93 & 11.66 & 14.64 & 0.70 & 25.98 & 9.67 & 47.37 & 6.93 \\
VoteNet+MCAN 	 &  18.15 & 48.56 & 29.63 & 17.80 & 11.57 & 7.10 & 29.12 & 11.68 & 53.34 & 10.36 \\
ScanRefer+MCAN (pipeline, real) & 16.47 & 49.05 & 18.71 & 10.97 & 16.53 & 0.76 & 22.45 & 8.76 & 40.81 & 6.41 \\
ScanRefer+MCAN (pipeline, mesh) & 16.39 & 47.76 & 18.28 & 10.42 & 15.03 & 0.71 & 22.07 & 8.62 & 40.19 & 6.03 \\
ScanRefer+MCAN (e2e) 	 &  19.04 & 49.70 & 26.98 & 16.17 & 11.28 & 7.82 & 28.61 & 11.38 & 53.41 & 10.63 \\
ScanQA (single) 	 & {20.90} & \textbf{54.11} & {30.68} & \textbf{21.20} & \textbf{15.81} & {10.75} & {31.09} & {12.59} & {60.24} & {11.29} \\
ScanQA (multiple)  &  {21.30} & 53.05 & \textbf{31.14} & \textbf{21.20} & \textbf{15.81} & \textbf{11.18} & \textbf{31.62} & \textbf{12.82} & \textbf{60.95} & \textbf{11.68} \\
\bottomrule
	\end{tabular}
    \caption{
        Performance comparison for question answering with image captioning metrics
    }
    \label{table:additional_overall_performance}

\end{center}
\end{table*}

%% file: supp_result/multi_module_ablation.tex
\begin{table*}[t]
\begin{center}
	\footnotesize\begin{tabular}{ccccccccccccc}
        \toprule
ANS & OBJ & LOC & EM@1 & EM@10 & BLEU-1 & BLEU-2 & BLEU-3 & BLEU-4 & ROUGE & METEOR & CIDEr & SPICE \\
\midrule
\multicolumn{3}{l}{\textbf{Valid}} \\
\checkmark & & 	 &  7.53 & 27.70 & 8.30 & 6.03 & 0.15 & 0.02 & 10.14 & 4.47 & 21.04 & 2.21 \\
\checkmark & \checkmark & 	 &  19.87 & 49.43 & 29.26 & 18.75 & 13.03 & 7.30 & 31.98 & 12.28 & 60.63 & 11.98 \\
\checkmark & & \checkmark 	 &  20.47 & 50.05 & 28.34 & 19.03 & 14.16 & 9.95 & 31.91 & 12.25 & 61.02 & 11.69 \\
\checkmark & \checkmark & \checkmark 	 &  \textbf{21.05} & \textbf{51.23} & \textbf{30.24} & \textbf{20.40} & \textbf{15.11} & \textbf{10.08} & \textbf{33.33} & \textbf{13.14} & \textbf{64.86} & \textbf{13.43} \\
\midrule
\multicolumn{3}{l}{\textbf{Test w/ objects}} \\
\checkmark & & 	 &  8.20 & 32.68 & 8.10 & 5.57 & 0.14 & 0.02 & 10.09 & 4.29 & 21.02 & 1.89 \\
\checkmark & \checkmark & 	 &  22.07 & 55.10 & 30.44 & 19.85 & 14.33 & 10.18 & 32.89 & 13.06 & 63.93 & 11.51 \\
\checkmark & & \checkmark 	 &  \textbf{23.77} & 55.31 & 30.65 & \textbf{21.42} & \textbf{16.61} & \textbf{13.37} & \textbf{34.15} & 13.40 & 66.69 & 10.88 \\
\checkmark & \checkmark & \checkmark 	 &  23.05 & \textbf{55.99} & \textbf{31.40} & 21.18 & 15.82 & 11.70 & 34.05 & \textbf{13.60} & \textbf{66.76} & \textbf{12.30} \\
\midrule
\multicolumn{3}{l}{\textbf{Test w/o objects}} \\
\checkmark & & 	 &  8.41 & 31.42 & 8.40 & 5.48 & 0.14 & 0.02 & 10.32 & 4.35 & 20.64 & 1.68 \\
\checkmark & \checkmark & 	 &  20.80 & 53.70 & \textbf{31.30} & 20.51 & 14.58 & 10.48 & 31.51 & 12.75 & 60.09 & \textbf{11.75} \\
\checkmark & & \checkmark 	 &  20.90 & \textbf{53.78} & 30.01 & \textbf{22.39} & \textbf{17.98} & \textbf{13.79} & 30.69 & 12.51 & 60.37 & 11.34 \\
\checkmark & \checkmark & \checkmark 	 &  \textbf{21.30} & 53.05 & 31.14 & 21.20 & 15.81 & 11.18 & \textbf{31.62} & \textbf{12.82} & \textbf{60.95} & 11.68 \\
\bottomrule
	\end{tabular}
    \caption{
        Performance comparison between the different experimental conditions of the ScanQA (multiple) model
    }
    \label{table:scanqa_multi_component_ablation}
\end{center}
\vspace{-4mm}
\end{table*}

%% file: supp_result/num_layer.tex
\begin{table*}[t]
\begin{center}
	\footnotesize\begin{tabular}{lcccccccccc}
        \toprule
Model                & EM@1 & EM@10 & BLEU-1 & BLEU-2 & BLEU-3 & BLEU-4 & ROUGE & METEOR & CIDEr & SPICE \\
\midrule
\textbf{Valid} \\
ScanQA ($L=1$) 	 &  19.96 & 49.78 & \textbf{29.49} & 19.16 & 13.23 & 8.44 & 32.35 & 12.59 & 61.14 & \textbf{12.61} \\
ScanQA ($L=2$) 	 &  \textbf{20.28} & \textbf{50.01} & 29.47 & \textbf{19.84} & \textbf{14.65} & \textbf{9.55} & \textbf{32.37} & \textbf{12.60} & \textbf{61.66} & 11.86 \\
ScanQA ($L=3$) 	 &  11.94 & 39.14 & 13.93 & 8.82 & 7.23 & 0.00 & 18.64 & 6.98 & 33.30 & 6.55 \\
\midrule
\textbf{Test w/ objects} \\
ScanQA ($L=1$) 	 &  \textbf{23.83} & 55.63 & \textbf{32.64} & \textbf{21.80} & 15.63 & 11.67 & \textbf{35.20} & \textbf{14.15} & \textbf{69.70} & \textbf{12.65} \\
ScanQA ($L=2$) 	 &  23.45 & \textbf{56.51} & 31.56 & 21.39 & \textbf{15.87} & \textbf{12.04} & 34.34 & 13.55 & 67.29 & 11.99 \\
ScanQA ($L=3$) 	 &  13.83 & 44.71 & 15.05 & 7.90 & 5.84 & 0.00 & 19.62 & 7.34 & 36.38 & 5.62 \\
\midrule
\textbf{Test w/o objects} \\
ScanQA ($L=1$) 	 &  \textbf{21.01} & 52.50 & \textbf{31.23} & \textbf{21.37} & \textbf{15.97} & \textbf{11.20} & \textbf{31.55} & \textbf{12.84} & \textbf{61.11} & \textbf{11.82} \\
ScanQA ($L=2$) 	 &  20.90 & \textbf{54.11} & 30.68 & 21.20 & 15.81 & 10.75 & 31.09 & 12.59 & 60.24 & 11.29 \\
ScanQA ($L=3$) 	 &  11.63 & 40.30 & 13.42 & 7.75 & 5.96 & 0.00 & 16.75 & 6.43 & 29.86 & 5.13 \\
\bottomrule
	\end{tabular}
    \caption{
        Performance comparison for the ScanQA model with difference number of layers $L$
    }
    \label{table:num_layer}
\end{center}
\vspace{-4mm}
\end{table*}

%% file: supp_result/hidden_size.tex
\begin{table*}[t]
\begin{center}
	\footnotesize\begin{tabular}{lcccccccccc}
        \toprule
Model                & EM@1 & EM@10 & BLEU-1 & BLEU-2 & BLEU-3 & BLEU-4 & ROUGE & METEOR & CIDEr & SPICE \\
\midrule
\textbf{Valid} \\
ScanQA ($d=128$) 	 &  \textbf{20.88} & \textbf{50.70} & \textbf{30.08} & \textbf{20.62} & \textbf{15.72} & \textbf{11.18} & \textbf{33.25} & \textbf{12.97} & \textbf{64.09} & \textbf{12.77} \\
ScanQA ($d=256$) 	 &  20.28 & 50.01 & 29.47 & 19.84 & 14.65 & 9.55 & 32.37 & 12.60 & 61.66 & 11.86 \\
ScanQA ($d=512$) 	 &  13.99 & 41.54 & 17.01 & 11.02 & 8.26 & 0.00 & 21.97 & 8.11 & 38.78 & 6.89 \\
\midrule
\textbf{Test w/ objects} \\
ScanQA ($d=128$) 	 &  \textbf{24.38} & \textbf{56.71} & \textbf{32.30} & \textbf{22.47} & \textbf{17.98} & \textbf{14.96} & \textbf{35.24} & \textbf{14.07} & \textbf{69.53} & \textbf{12.61} \\
ScanQA ($d=256$) 	 &  23.45 & 56.51 & 31.56 & 21.39 & 15.87 & 12.04 & 34.34 & 13.55 & 67.29 & 11.99 \\
ScanQA ($d=512$) 	 &  14.75 & 46.02 & 16.89 & 9.18 & 7.06 & 6.77 & 21.17 & 7.76 & 38.56 & 5.47 \\
\midrule
\textbf{Test w/o objects} \\
ScanQA ($d=128$) 	 &  \textbf{21.17} & \textbf{54.20} & \textbf{31.79} & \textbf{22.23} & \textbf{16.65} & \textbf{11.32} & \textbf{31.83} & \textbf{13.01} & \textbf{61.92} & \textbf{12.13} \\
ScanQA ($d=256$) 	 &  20.90 & 54.11 & 30.68 & 21.20 & 15.81 & 10.75 & 31.09 & 12.59 & 60.24 & 11.29 \\
ScanQA ($d=512$) 	 &  12.83 & 43.28 & 16.67 & 9.56 & 7.01 & 4.70 & 18.97 & 7.31 & 33.63 & 5.08 \\
\bottomrule
	\end{tabular}
    \caption{
        Performance comparison for the ScanQA model with difference hidden size $d$
    }
    \label{table:hidden_size}
\end{center}
\vspace{-4mm}
\end{table*}

%% file: supp_result/question_type.tex
\subsection{Accuracy for Question Types}

\begin{table*}[t]
\begin{center}
	\footnotesize\begin{tabular}{llc}
        \toprule
Question type & Question beginning & \# Instances in the valid set \\
\midrule\midrule
Object & \textit{What is}~~(except questions classified as color) & 1476 \\
\midrule
Color & \textit{What color}  & 838 \\
      & \textit{What is the color} \\
\midrule
Object nature & \textit{What type} & 358 \\
 & \textit{What shape} \\
 & \textit{What kind} \\
\midrule
Place & \textit{Where is} & 963 \\
\midrule
Number & \textit{How many} & 224 \\
\midrule
Other & (remaining questions) & 814 \\
\bottomrule
\end{tabular}
    \caption{
        Question type and the beginning of the question sentence
    }
    \label{table:define_question_type}
\end{center}
\vspace{-4mm}
\end{table*}

We classify the questions into six types: \textit{object}, \textit{color}, \textit{object nature}, \textit{place}, \textit{number} and \textit{other} by the beginning of the question sentences following Table~\ref{table:define_question_type}.
We evaluate the detailed question answering performance for each class for 2D-QA baselines and the ScanQA model.
Table~\ref{table:performance3D_question_time_valid},
Table~\ref{table:performance3D_question_time_test_w_object} and
Table~\ref{table:performance3D_question_time_test_wo_object} presents the detailed accuracy on valid, test w/ object and test w/o object respectively.

We notice that the performance scores for \textit{place} questions is lower than other questions.
We assume there are two reasons for this.
First, there are several ways to answer the questions of \textit{place}. Model predicts longer answers and therefor image captioning-based metrics are suitable rather than exact matching for \textit{place} questions.
For \textit{color} questions, possible answers are limited and simple 2D-QA model can answer the questions from the sights of objects from several images. However, such answers are not grounded to objects in the questions.

\begin{table*}[t]
\begin{center}
	\scriptsize
	\begin{tabular}{lcccccccccc}
        \toprule
Model                & EM@1 & EM@10 & BLEU-1 & BLEU-2 & BLEU-3 & BLEU-4 & ROUGE & METEOR & CIDEr & SPICE \\
\midrule
\textbf{Object} \\
RandomImage+MCAN & 15.02 & 45.06 & 20.33 & 16.22 & \textbf{17.13} & \textbf{0.79} & 43.03 & 9.18 & 22.18 & 14.03 \\
VoteNet+MCAN & 12.31 & 41.07 & 20.81 & 13.51 & 6.47 & 0.00 & 39.81 & 8.88 & 21.37 & 14.09 \\
ScanRefer+MCAN (pipeline) & 11.84 & 38.09 & 15.91 & 12.73 & 0.23 & 0.03 & 32.64 & 7.37 & 17.49 & 12.40 \\
ScanRefer+MCAN (e2e) & 14.82 & 42.76 & 20.43 & 15.57 & 8.19 & 0.00 & 41.74 & 9.00 & 22.12 & 14.87 \\
ScanQA (single) & 17.52 & 45.47 & 23.94 & 18.19 & 0.00 & 0.00 & 50.05 & 10.62 & 26.01 & 17.57 \\
ScanQA (multiple) & \textbf{18.27} & \textbf{47.70} & \textbf{26.15} & \textbf{19.19} & 14.46 & 0.00 & \textbf{53.84} & \textbf{11.53} & \textbf{27.52} & \textbf{18.46} \\
\midrule
\textbf{Color} \\
RandomImage+MCAN & 44.03 & \textbf{86.04} & 45.92 & 31.38 & 0.44 & \textbf{0.05} & 86.65 & 23.57 & 49.36 & 2.19 \\
VoteNet+MCAN & 41.65 & 81.62 & \textbf{47.01} & 28.19 & \textbf{20.58} & 0.01 & 86.54 & 23.32 & 49.01 & \textbf{3.09} \\
ScanRefer+MCAN (pipeline) & 30.79 & 83.41 & 32.75 & 0.05 & 0.01 & 0.00 & 59.01 & 16.14 & 34.87 & 0.00 \\
ScanRefer+MCAN (e2e) & \textbf{44.99} & 83.77 & 46.72 & \textbf{35.91} & 0.00 & 0.00 & \textbf{87.75} & \textbf{24.16} & \textbf{50.13} & 1.85 \\
ScanQA (single) & 42.60 & 83.53 & 43.92 & 29.48 & 0.00 & 0.00 & 84.42 & 22.61 & 47.68 & 1.39 \\
ScanQA (multiple) & 42.60 & 85.44 & 45.23 & 17.78 & 0.00 & 0.00 & 85.20 & 22.76 & 48.34 & 2.41 \\
\midrule
\textbf{Object nature} \\
RandomImage+MCAN & 18.16 & 47.49 & 31.06 & 26.40 & \textbf{26.83} & \textbf{1.07} & 57.15 & 13.18 & 33.69 & 7.11 \\
VoteNet+MCAN & 17.04 & 47.77 & 35.70 & 23.15 & 16.71 & 0.00 & 62.83 & 14.66 & 35.21 & 16.42 \\
ScanRefer+MCAN (pipeline) & 15.36 & 43.30 & 20.14 & 16.72 & 18.38 & 0.77 & 38.60 & 9.25 & 27.09 & 2.27 \\
ScanRefer+MCAN (e2e) & 13.41 & 48.32 & 36.50 & 20.03 & 16.41 & 0.00 & 58.03 & 14.50 & 34.75 & 16.08 \\
ScanQA (single) & 18.44 & 51.40 & \textbf{41.65} & 25.65 & 18.81 & 0.00 & 73.26 & 16.54 & \textbf{41.61} & 17.16 \\
ScanQA (multiple) & \textbf{20.11} & \textbf{54.75} & 41.34 & \textbf{29.72} & 25.80 & 0.01 & \textbf{78.31} & \textbf{17.31} & 40.54 & \textbf{18.67} \\
\midrule
\textbf{Place} \\
RandomImage+MCAN & 3.95 & 17.24 & 17.87 & 11.59 & 8.29 & 0.00 & 33.58 & 8.05 & 19.00 & 11.55 \\
VoteNet+MCAN & 4.88 & 18.48 & 28.31 & 17.84 & 12.07 & 7.47 & 45.08 & 9.97 & 26.41 & 14.68 \\
ScanRefer+MCAN (pipeline) & 2.08 & 11.53 & 8.26 & 5.08 & 0.11 & 0.02 & 14.34 & 5.29 & 9.65 & 5.74 \\
ScanRefer+MCAN (e2e) & 4.98 & 17.86 & 25.31 & 15.64 & 11.20 & 8.01 & 44.72 & 9.87 & 25.06 & 15.98 \\
ScanQA (single) & 6.85 & 23.16 & \textbf{28.78} & \textbf{19.48} & \textbf{14.41} & 9.55 & 57.00 & \textbf{11.49} & 28.19 & 16.66 \\
ScanQA (multiple) & \textbf{7.79} & \textbf{23.99} & 28.21 & 19.13 & 14.27 & \textbf{9.92} & \textbf{59.34} & 11.46 & \textbf{28.30} & \textbf{18.13} \\
\midrule
\textbf{Number} \\
RandomImage+MCAN & 39.29 & \textbf{86.16} & 45.01 & 0.06 & 0.01 & 0.00 & 72.26 & 19.55 & 46.70 & 0.00 \\
VoteNet+MCAN & 36.16 & \textbf{86.16} & 43.77 & 20.59 & 0.33 & 0.04 & 67.56 & 18.13 & 44.02 & 0.40 \\
ScanRefer+MCAN (pipeline) & 38.39 & 85.71 & 44.26 & 0.06 & 0.01 & 0.00 & 67.76 & 19.37 & 45.69 & 0.00 \\
ScanRefer+MCAN (e2e) & 34.82 & 84.82 & 39.01 & 0.06 & 0.01 & 0.00 & 61.06 & 17.01 & 40.78 & 0.00 \\
ScanQA (single) & 39.29 & 85.71 & 44.29 & 0.00 & 0.00 & 0.00 & 72.15 & 19.16 & 46.05 & 0.00 \\
ScanQA (multiple) & \textbf{40.62} & 85.71 & \textbf{46.00} & \textbf{44.09} & \textbf{0.55} & \textbf{0.06} & \textbf{75.68} & \textbf{19.80} & \textbf{47.72} & \textbf{0.45} \\
\midrule
\textbf{Other} \\
RandomImage+MCAN & 14.13 & 41.15 & 18.93 & 13.19 & \textbf{11.63} & \textbf{0.54} & 42.28 & 8.83 & 24.75 & 9.17 \\
VoteNet+MCAN & 11.06 & 36.36 & 20.87 & 7.91 & 0.00 & 0.00 & 36.71 & 8.61 & 23.29 & 7.68 \\
ScanRefer+MCAN (pipeline) & 10.69 & 37.22 & 17.28 & 10.45 & 0.18 & 0.02 & 33.89 & 8.12 & 21.74 & 6.98 \\
ScanRefer+MCAN (e2e) & 12.16 & 38.94 & 21.10 & 12.82 & 8.29 & 0.00 & 42.69 & 9.32 & 24.54 & 9.91 \\
ScanQA (single) & 14.13 & 43.37 & 22.26 & 13.72 & 8.02 & 0.00 & 45.39 & 9.96 & 26.30 & 10.78 \\
ScanQA (multiple) & \textbf{14.62} & \textbf{43.61} & \textbf{23.52} & \textbf{15.64} & 9.61 & 0.00 & \textbf{47.89} & \textbf{10.39} & \textbf{27.26} & \textbf{11.34} \\
\bottomrule
	\end{tabular}
    \caption{
        Valid set performance comparison for question answering with image captioning metrics. \textbf{e2e} represents an end-to-end model.
    }
    \label{table:performance3D_question_time_valid}
\end{center}
\vspace{-4mm}
\end{table*}

\begin{table*}[t]
\begin{center}
	\scriptsize
	\begin{tabular}{lcccccccccc}
        \toprule
Model                & EM@1 & EM@10 & BLEU-1 & BLEU-2 & BLEU-3 & BLEU-4 & ROUGE & METEOR & CIDEr & SPICE \\
\midrule
\textbf{Object} \\
RandomImage+MCAN & 16.47 & 46.59 & 22.68 & 18.15 & 0.00 & 0.00 & 47.16 & 9.69 & 23.49 & 15.65 \\
VoteNet+MCAN & 13.23 & 43.84 & 21.32 & 13.72 & 7.47 & 5.36 & 40.98 & 8.80 & 21.05 & 14.25 \\
ScanRefer+MCAN (pipeline) & 12.74 & 41.24 & 15.87 & 12.36 & 0.23 & 0.03 & 33.82 & 7.10 & 17.32 & 13.24 \\
ScanRefer+MCAN (e2e) & 15.97 & 46.24 & 21.12 & 15.76 & 6.96 & 0.00 & 43.01 & 9.13 & 22.28 & 15.87 \\
ScanQA (single) & 18.86 & 49.96 & 24.75 & 18.78 & 13.73 & 10.62 & 51.81 & 10.68 & 26.00 & 17.09 \\
ScanQA (multiple) & \textbf{19.14} & \textbf{52.15} & \textbf{26.78} & \textbf{20.97} & \textbf{17.86} & \textbf{17.64} & \textbf{55.24} & \textbf{11.25} & \textbf{26.98} & \textbf{18.47} \\
\midrule
\textbf{Color} \\
RandomImage+MCAN & 45.56 & \textbf{89.56} & 49.39 & \textbf{50.81} & 0.62 & \textbf{0.07} & 91.84 & 25.87 & 50.08 & 2.08 \\
VoteNet+MCAN & 40.57 & 84.66 & 48.05 & 23.43 & \textbf{11.85} & 0.00 & 85.40 & 23.56 & 47.41 & \textbf{2.28} \\
ScanRefer+MCAN (pipeline) & 31.24 & 86.69 & 36.03 & 0.06 & 0.01 & 0.00 & 62.00 & 17.61 & 35.67 & 0.00 \\
ScanRefer+MCAN (e2e) & 41.96 & 88.08 & 47.76 & 27.33 & 0.00 & 0.00 & 84.52 & 24.04 & 47.74 & 2.22 \\
ScanQA (single) & \textbf{45.75} & 88.45 & \textbf{50.27} & 22.71 & 0.00 & 0.00 & \textbf{92.54} & \textbf{26.06} & \textbf{50.96} & 2.10 \\
ScanQA (multiple) & 43.25 & 88.54 & 48.71 & 12.31 & 0.00 & 0.00 & 88.00 & 24.53 & 48.83 & 2.14 \\
\midrule
\textbf{Object nature} \\
RandomImage+MCAN & 19.91 & 50.98 & 35.93 & 25.12 & \textbf{23.14} & \textbf{0.98} & 56.79 & 13.73 & 35.93 & 6.54 \\
VoteNet+MCAN & 19.91 & 49.45 & 38.95 & 22.65 & 0.00 & 0.00 & 61.17 & 14.90 & 36.42 & \textbf{18.18} \\
ScanRefer+MCAN (pipeline) & 19.69 & 46.17 & 27.85 & 23.18 & 0.00 & 0.00 & 47.31 & 11.42 & 33.08 & 2.33 \\
ScanRefer+MCAN (e2e) & 19.04 & 50.33 & 39.23 & 23.34 & 17.01 & 0.00 & 61.06 & 14.71 & 35.99 & 15.24 \\
ScanQA (single) & \textbf{22.98} & \textbf{56.67} & 45.09 & 28.98 & 21.27 & 0.96 & \textbf{72.81} & 16.85 & 41.52 & 14.79 \\
ScanQA (multiple) & 22.76 & 55.14 & \textbf{45.76} & \textbf{29.72} & 18.14 & 0.00 & 72.16 & \textbf{17.15} & \textbf{41.77} & 16.95 \\
\midrule
\textbf{Place} \\
RandomImage+MCAN & 3.97 & 16.10 & 19.50 & 14.32 & 13.07 & 11.67 & 39.74 & 8.68 & 19.87 & 10.97 \\
VoteNet+MCAN & 4.43 & 17.62 & 28.82 & 18.10 & 11.84 & 7.05 & 46.59 & 10.31 & 26.70 & 15.02 \\
ScanRefer+MCAN (pipeline) & 1.40 & 9.68 & 7.33 & 3.65 & 0.09 & 0.01 & 11.45 & 5.11 & 8.52 & 4.45 \\
ScanRefer+MCAN (e2e) & 4.08 & 18.20 & 24.31 & 15.62 & 11.18 & 7.35 & 43.72 & 9.64 & 24.54 & 14.32 \\
ScanQA (single) & \textbf{7.12} & \textbf{22.17} & \textbf{29.59} & \textbf{20.93} & \textbf{16.47} & \textbf{12.59} & \textbf{61.89} & \textbf{12.06} & \textbf{29.35} & \textbf{18.17} \\
ScanQA (multiple) & 5.95 & 21.94 & 26.89 & 18.87 & 14.53 & 10.60 & 55.13 & 11.11 & 27.10 & 16.74 \\
\midrule
\textbf{Number} \\
RandomImage+MCAN & \textbf{43.14} & 90.20 & 44.97 & 0.06 & 0.01 & 0.00 & \textbf{83.02} & 23.81 & 46.94 & 0.00 \\
VoteNet+MCAN & 39.61 & \textbf{90.98} & 44.01 & 16.07 & 0.29 & 0.04 & 75.94 & 21.71 & 44.29 & \textbf{0.72} \\
ScanRefer+MCAN (pipeline) & 43.53 & 89.41 & \textbf{45.32} & 0.06 & 0.01 & 0.00 & 81.52 & \textbf{23.89} & \textbf{47.48} & 0.00 \\
ScanRefer+MCAN (e2e) & 34.12 & 87.84 & 36.29 & 0.00 & 0.00 & 0.00 & 63.70 & 18.28 & 37.43 & 0.00 \\
ScanQA (single) & 40.78 & 90.59 & 44.94 & 0.00 & 0.00 & 0.00 & 77.33 & 22.44 & 45.80 & 0.00 \\
ScanQA (multiple) & 37.25 & 90.59 & 40.25 & \textbf{22.86} & \textbf{0.36} & \textbf{0.05} & 72.15 & 20.07 & 41.37 & 0.39 \\
\midrule
\textbf{Other} \\
RandomImage+MCAN & 16.37 & 45.46 & 19.02 & 12.93 & 0.00 & 0.00 & 42.83 & 9.01 & 24.99 & 9.02 \\
VoteNet+MCAN & 13.72 & 41.81 & 20.85 & 9.92 & 3.53 & 0.00 & 41.08 & 8.81 & 24.41 & 8.72 \\
ScanRefer+MCAN (pipeline) & 15.04 & 42.48 & 18.00 & 10.61 & 0.19 & 0.03 & 37.78 & 8.51 & 23.46 & 6.42 \\
ScanRefer+MCAN (e2e) & 14.71 & 42.59 & 21.09 & 11.01 & 7.21 & 0.00 & 41.46 & 9.23 & 24.70 & 9.36 \\
ScanQA (single) & \textbf{17.92} & \textbf{46.79} & \textbf{24.50} & \textbf{15.76} & \textbf{9.92} & \textbf{7.06} & \textbf{50.88} & \textbf{10.66} & \textbf{28.78} & \textbf{11.69} \\
ScanQA (multiple) & 17.37 & 46.02 & 23.94 & 14.30 & 8.00 & 0.00 & 50.63 & 10.55 & 28.06 & 11.58 \\
\bottomrule
	\end{tabular}
    \caption{
        Test w/ object set performance comparison for question answering with image captioning metrics. \textbf{e2e} represents an end-to-end model.
    }
    \label{table:performance3D_question_time_test_w_object}
\end{center}
\vspace{-4mm}
\end{table*}

\begin{table*}[t]
\begin{center}
	\scriptsize
	\begin{tabular}{lcccccccccc}
        \toprule
Model                & EM@1 & EM@10 & BLEU-1 & BLEU-2 & BLEU-3 & BLEU-4 & ROUGE & METEOR & CIDEr & SPICE \\
\midrule
\textbf{Object} \\
RandomImage+MCAN & 15.85 & 44.84 & 22.37 & 17.97 & 15.07 & \textbf{0.73} & 41.56 & 9.29 & 22.39 & 14.41 \\
VoteNet+MCAN & 13.80 & 42.52 & 20.65 & 12.39 & 5.86 & 0.00 & 37.06 & 8.71 & 20.69 & 14.79 \\
ScanRefer+MCAN (pipeline) & 12.48 & 41.14 & 17.70 & 14.87 & 0.27 & 0.04 & 32.31 & 7.44 & 17.52 & 12.95 \\
ScanRefer+MCAN (e2e) & 14.30 & 44.56 & 20.26 & 15.31 & 9.77 & 0.00 & 37.90 & 8.67 & 20.56 & 15.61 \\
ScanQA (single) & 17.39 & \textbf{49.75} & 24.79 & 18.87 & 11.47 & 0.00 & 46.97 & 10.36 & 24.89 & 16.73 \\
ScanQA (multiple) & \textbf{18.83} & 48.48 & \textbf{25.96} & \textbf{21.26} & \textbf{16.90} & 0.00 & \textbf{49.72} & \textbf{11.00} & \textbf{25.95} & \textbf{17.00} \\
\midrule
\textbf{Color} \\
RandomImage+MCAN & \textbf{43.31} & \textbf{89.38} & \textbf{45.74} & 0.00 & 0.00 & 0.00 & \textbf{87.49} & \textbf{24.55} & \textbf{46.66} & 0.90 \\
VoteNet+MCAN & 38.62 & 83.00 & 44.74 & \textbf{15.35} & \textbf{11.79} & \textbf{0.63} & 79.21 & 22.16 & 44.35 & \textbf{1.03} \\
ScanRefer+MCAN (pipeline) & 33.46 & 87.77 & 36.88 & 0.06 & 0.01 & 0.00 & 67.21 & 18.88 & 36.90 & 0.00 \\
ScanRefer+MCAN (e2e) & 40.15 & 85.23 & 44.53 & 0.00 & 0.00 & 0.00 & 81.64 & 22.92 & 44.72 & 0.71 \\
ScanQA (single) & 40.00 & 87.85 & 43.58 & 0.00 & 0.00 & 0.00 & 80.77 & 22.60 & 43.89 & 0.54 \\
ScanQA (multiple) & 40.92 & 87.46 & 44.15 & 14.64 & 0.00 & 0.00 & 82.22 & 22.96 & 44.62 & 0.77 \\
\midrule
\textbf{Object nature} \\
RandomImage+MCAN & 15.17 & 41.67 & 29.06 & 25.84 & 23.07 & 0.98 & 47.21 & 11.46 & 27.54 & 7.27 \\
VoteNet+MCAN & 10.68 & 41.45 & 33.29 & 18.02 & 11.91 & 0.00 & 45.85 & 12.37 & 28.44 & 15.82 \\
ScanRefer+MCAN (pipeline) & 13.46 & 39.74 & 20.73 & 22.67 & \textbf{25.65} & \textbf{1.03} & 35.67 & 8.29 & 22.18 & 3.46 \\
ScanRefer+MCAN (e2e) & 13.25 & 39.32 & 34.51 & 19.23 & 13.25 & 0.00 & 51.52 & 12.96 & 30.13 & 18.28 \\
ScanQA (single) & \textbf{15.60} & 46.15 & \textbf{41.96} & \textbf{28.75} & 19.96 & 0.92 & \textbf{65.48} & \textbf{15.62} & \textbf{35.60} & \textbf{20.89} \\
ScanQA (multiple) & 13.68 & \textbf{47.65} & 41.40 & 25.58 & 20.39 & 0.01 & 59.70 & 14.75 & 34.43 & 19.06 \\
\midrule
\textbf{Place} \\
RandomImage+MCAN & 4.76 & 19.34 & 21.88 & 14.48 & 11.08 & 7.76 & 41.12 & 9.18 & 22.08 & 12.85 \\
VoteNet+MCAN & 5.85 & 20.20 & 32.15 & 20.62 & 13.86 & 8.80 & 50.54 & 11.16 & 29.06 & 16.43 \\
ScanRefer+MCAN (pipeline) & 1.64 & 11.93 & 8.87 & 4.58 & 0.11 & 0.02 & 12.82 & 5.34 & 9.14 & 5.42 \\
ScanRefer+MCAN (e2e) & 4.99 & 21.14 & 24.47 & 15.03 & 10.76 & 7.95 & 44.46 & 9.57 & 23.99 & 15.05 \\
ScanQA (single) & \textbf{6.79} & \textbf{26.37} & \textbf{32.47} & \textbf{23.04} & \textbf{18.34} & \textbf{13.74} & \textbf{63.40} & \textbf{12.27} & \textbf{29.67} & \textbf{19.28} \\
ScanQA (multiple) & \textbf{6.79} & 24.57 & 30.18 & 20.39 & 15.19 & 10.81 & 56.65 & 11.37 & 27.97 & 17.71 \\
\midrule
\textbf{Number} \\
RandomImage+MCAN & \textbf{42.51} & \textbf{90.64} & 45.13 & 0.07 & 0.01 & 0.00 & \textbf{77.60} & \textbf{22.40} & \textbf{45.21} & 0.00 \\
VoteNet+MCAN & 33.69 & \textbf{90.64} & 40.35 & 25.05 & 0.40 & \textbf{0.05} & 63.81 & 18.53 & 38.63 & \textbf{0.27} \\
ScanRefer+MCAN (pipeline) & 41.44 & \textbf{90.64} & 44.12 & 0.06 & 0.01 & 0.00 & 74.31 & 22.01 & 44.17 & 0.00 \\
ScanRefer+MCAN (e2e) & 37.97 & 89.30 & 41.12 & 0.00 & 0.00 & 0.00 & 70.08 & 19.69 & 40.83 & 0.00 \\
ScanQA (single) & 36.90 & \textbf{90.64} & 42.69 & 28.69 & 0.43 & \textbf{0.05} & 68.30 & 19.57 & 41.43 & 0.00 \\
ScanQA (multiple) & 40.91 & \textbf{90.64} & \textbf{46.20} & \textbf{29.85} & \textbf{0.44} & \textbf{0.05} & 77.26 & 21.84 & 45.16 & 0.00 \\
\midrule
\textbf{Other} \\
RandomImage+MCAN & 15.21 & \textbf{43.11} & 18.74 & \textbf{14.85} & \textbf{15.93} & 0.72 & \textbf{40.50} & 8.65 & 22.32 & 8.13 \\
VoteNet+MCAN & 12.36 & 37.75 & 20.15 & 9.26 & 3.22 & 0.00 & 33.82 & 8.25 & 20.73 & 7.64 \\
ScanRefer+MCAN (pipeline) & 11.82 & 40.81 & 16.74 & 11.26 & 0.00 & 0.00 & 32.49 & 7.23 & 19.04 & 5.50 \\
ScanRefer+MCAN (e2e) & 13.35 & 38.51 & 20.31 & 9.82 & 6.21 & 0.00 & 37.70 & 8.73 & 22.34 & 9.12 \\
ScanQA (single) & \textbf{15.32} & 41.79 & \textbf{21.99} & 12.74 & 6.92 & 0.00 & 40.30 & \textbf{9.30} & \textbf{23.99} & \textbf{10.33} \\
ScanQA (multiple) & 14.55 & 40.48 & 20.69 & 12.93 & 8.57 & \textbf{6.81} & 38.74 & 8.69 & 22.53 & 9.22 \\
\bottomrule
	\end{tabular}
    \caption{
        Test w/o object set performance comparison for question answering with image captioning metrics. \textbf{e2e} represents an end-to-end model.
    }
    \label{table:performance3D_question_time_test_wo_object}
\end{center}
\vspace{-4mm}
\end{table*}